\documentclass{article} %
\usepackage{iclr2027_conference,times}

\usepackage{amsmath,amsfonts,bm}

\def\eqref#1{equation~\ref{#1}}

\def\1{\bm{1}}

\DeclareMathAlphabet{\mathsfit}{\encodingdefault}{\sfdefault}{m}{sl}
\SetMathAlphabet{\mathsfit}{bold}{\encodingdefault}{\sfdefault}{bx}{n}

\usepackage{amsmath}
\usepackage{amssymb}
\usepackage{booktabs}
\usepackage{graphicx}
\usepackage{microtype}
\usepackage{hyperref}
\usepackage{url}
\providecommand{\tbd}{\textsf{TBD}}
\title{GLoC-EHR: Evidence-Cited Clinical Reasoning\\
over Global Context and Local EHR Events}

\author{Chaiho Shin$^{1}$ \quad Kwangsoo Kim$^{2,3,4}$\thanks{Corresponding author.} \\
{\small $^{1}$Interdisciplinary Program of Medical Informatics, Seoul National University College of Medicine} \\
{\small $^{2}$Department of Transdisciplinary Medicine, Seoul National University Hospital} \\
{\small $^{3}$Center for Data Science, Healthcare AI Research Institute, Seoul National University Hospital} \\
{\small $^{4}$Department of Medicine, Seoul National University College of Medicine}
}

\iclrfinalcopy
\begin{document}

\setcounter{footnote}{1}
\maketitle
\lhead{Preprint}

\begin{abstract}
Structured electronic health records (EHRs) contain a patient's clinical trajectory as a sequence of clinical codes.
Answering clinical questions from such records requires both the context of the whole trajectory and the specific events that support the answer.
We introduce GLoC-EHR, a multimodal language model that reads a contextual encoding of the record through a fixed-size global memory of the trajectory and a local memory of selected events.
The model learns to generate hospital-course summaries from the global memory and descriptions of masked concepts from the local memory, aligning both with clinical text.
It is then trained to cite evidence before answering, through rationale fine-tuning followed by group relative policy optimization (GRPO) with rewards for correct answers and record-supported evidence.
On three MIMIC-IV outcome tasks, GLoC-EHR attains the highest macro AUROC among the compared models when it answers directly, whereas zero-shot LLMs reading the serialized record fall far behind.
With evidence-cited reasoning, it stays close to its direct multi-task counterpart in macro AUROC, and the evidence terms of the objective reduce unsupported evidence at a similar macro AUROC.
The local memory adds distinct supported findings, particularly under strict matching, without a detectable change in macro AUROC.
Without retraining, GLoC-EHR transfers to EHRSHOT on par with EHR-BERT and answers two unseen laboratory questions better than zero-shot prompting of its own backbone.
\end{abstract}

\section{Introduction}
\label{sec:introduction}

Electronic health records (EHRs) capture diagnoses, medication orders,
procedures, and measurements over time, providing a rich longitudinal account
of a patient's health and care. These records offer both a broader view of the
patient's clinical trajectory and specific observations relevant to a clinical
question. However, clinically relevant information can be dispersed across lengthy, heterogeneous patient records, making it challenging for clinicians to identify and synthesize information needed to guide patient care \citep{reichert2010cognitive}.

To harness longitudinal EHR data for clinical prediction, EHR-based models have been developed
for a wide range of clinical
outcome-prediction tasks \citep{choi2016retain,li2020behrt,rasmy2021medbert}.
In particular, pretrained EHR encoders learn contextual representations of
medical codes from patient histories. BEHRT adapts masked language modeling to
diagnosis sequences with age and visit information \citep{li2020behrt}; Med-BERT
learns transferable diagnosis representations through large-scale pretraining
\citep{rasmy2021medbert}; and CEHR-BERT incorporates temporal information across
multiple structured domains \citep{pang2021cehrbert}. These models provide a
reusable foundation for prediction and can reduce the need for labeled data.
However, their usual downstream interface is a task-specific predictive head:
adapting to a new outcome generally requires fitting a new head or fine-tuning
the model with corresponding labels
\citep{rasmy2021medbert,wornow2023ehrshot}. Thus, transfer remains dependent on
task-specific supervision, even when the underlying representation is reused.
Moreover, an outcome score alone does not express the recorded findings that
support a clinical inference. Extending these representations to
natural-language reasoning therefore requires connecting the encoded record,
the clinical question, and the evidence supporting an answer.

One approach to connecting structured EHRs with language models, which encode
substantial clinical knowledge \citep{singhal2023medpalm,jiang2023nyutron}, is to
serialize medical codes into textual descriptions that the models can embed or
reason over \citep{hegselmann2023tabllm,shoham2024cpllm,zhu2024prompting,hegselmann2026ehrencoders,lin2025ehrmind,liao2025ehrr1}.
Serialization retains an explicit account of the selected events at a growing
context cost \citep{wornow2025contextclues}.
In our downstream cohorts, serializing even the 24-hour observation window
yields an average of 4.7k tokens (90th percentile: 14k), increasing to 12.8k
(48k) for the full index admission.
Other approaches use intermediate textual summaries \citep{gao2026moma} or
fuse language-model embeddings of domain-specific pseudo-notes
\citep{lee2025meme}. Such compression may omit details needed to
substantiate a clinical inference. A distinct approach,
Llemr, aligns individually encoded clinical events with a generative language
model \citep{wu2024llemr}, reducing the input length relative to textual
serialization. However, the resulting input still scales with the number of
events.

Connecting structured EHRs with language models makes clinical information
available for reasoning, but does not by itself ensure that the generated
reasoning is supported by the patient's record. Recent work, including EHRMIND
\citep{lin2025ehrmind} and EHR-R1 \citep{liao2025ehrr1}, develops EHR-based
reasoning through reinforcement learning with rewards for final-answer
correctness, which do not explicitly distinguish supported from unsupported
findings.

We therefore introduce GLoC-EHR, an EHR--text multimodal large language model
that combines a global--local interface with evidence-aware reward training.
A contextual EHR encoder reaches the language model through two routes: a global
route, whose compact memory summarizes the observed trajectory, and a local
route, whose memory retains selected individual events. Trajectory alignment,
which generates the hospital-course summary of an admission, aligns the global
route with clinical text, and concept alignment, which recovers descriptions of
masked concepts, aligns the local route. Beyond answer correctness, an
evidence-matching reward credits findings supported by the patient's record
rather than concepts merely associated with the task. We study how the local
route and this reward shape the generated reasoning on clinical outcome tasks, with direct prediction as a
reference for the information in the representation. Our contributions are
threefold:
\begin{itemize}
    \setlength{\itemsep}{0pt}\setlength{\parskip}{0pt}\setlength{\parsep}{0pt}
    \item A global--local interface whose two routes are aligned with clinical
    text, giving a language model trajectory context and access to individual
    events; answering directly, GLoC-EHR attains the highest macro AUROC over
    three MIMIC-IV tasks (Section~\ref{sec:outcome_comparison}).
    \item An evidence-aware objective that combines this reward with a
    sufficiency loss on the cited findings; it reduces unsupported evidence
    with no detectable change in macro AUROC, which stays close to that of
    direct prediction
    (Sections~\ref{sec:outcome_comparison}--\ref{sec:ablations}).
    \item An evaluation of cited evidence under stricter matching and cited-slot
    masking, showing that the local route shapes which findings are cited, with no
    detectable effect on the answer; without retraining, GLoC-EHR also matches EHR-BERT on
    EHRSHOT and outperforms its zero-shot backbone on unseen laboratory questions (Sections~\ref{sec:evidence} and~\ref{sec:external}).
\end{itemize}

\section{Related Work}
\label{sec:related_work}

\paragraph{Structured EHR representations.}
RETAIN uses reverse-time attention over visits and variables
\citep{choi2016retain}, and the BERT-style encoders discussed above learn
reusable contextual code representations
\citep{li2020behrt,rasmy2021medbert,pang2021cehrbert}. Autoregressive and
time-to-event foundation models are instead pretrained to forecast future events
in patient timelines
\citep{steinberg2021clmbr,steinberg2024motor,kraljevic2024foresight,renc2024ethos}.
We build on contextual representations to support language generation grounded
in the record.

\paragraph{EHR--language alignment.}
Clinical text connects structured records with language models in several
roles: serialized records read or embedded for prediction
\citep{hegselmann2023tabllm,shoham2024cpllm,hegselmann2026ehrencoders},
post-training for EHR reasoning, including reinforcement learning
\citep{lin2025ehrmind,liao2025ehrr1},
modality-specialist agents that summarize non-text inputs \citep{gao2026moma},
and pseudo-notes embedded by a frozen language model \citep{lee2025meme}.
Closest to our setting, Llemr connects a frozen clinical event encoder to an LLM
through a projection, as multimodal LLMs do for images, time series, and
individual health data \citep{liu2023llava,jin2024timellm,belyaeva2023helm}, and
trains it with schema alignment followed by reasoning instruction tuning
\citep{wu2024llemr}. Our interface adds a compressed trajectory memory to a
separate memory of selected events, and our reward checks whether the generated
findings are supported by the record.

\paragraph{Evidence support and rationale faithfulness.}
Attribution metrics check whether generated statements are supported by a source
\citep{rashkin2023ais,gao2023alce,min2023factscore}, but a rationale's
consistency with its input differs from its faithfulness to the process that
produced an answer \citep{jacovi2020faithfulness,turpin2023unfaithful}. ERASER
separates agreement with evidence annotations from behavioral tests of
comprehensiveness and sufficiency \citep{deyoung2020eraser}, and studies of
natural language explanations and chain-of-thought use interventions to test
whether stated reasons affect predictions
\citep{atanasova2023faithfulness,lanham2023faithfulness}. Accordingly, our
matching reward for group relative policy optimization (GRPO)
\citep{shao2024deepseekmath} measures support in the record rather than
causation, and we evaluate it together with evidence coverage and an
intervention that removes cited versus comparable uncited local representations
before regenerating the response.

\section{Method}
\label{sec:method}

\begin{figure}[t]
    \centering
    \includegraphics[width=\linewidth]{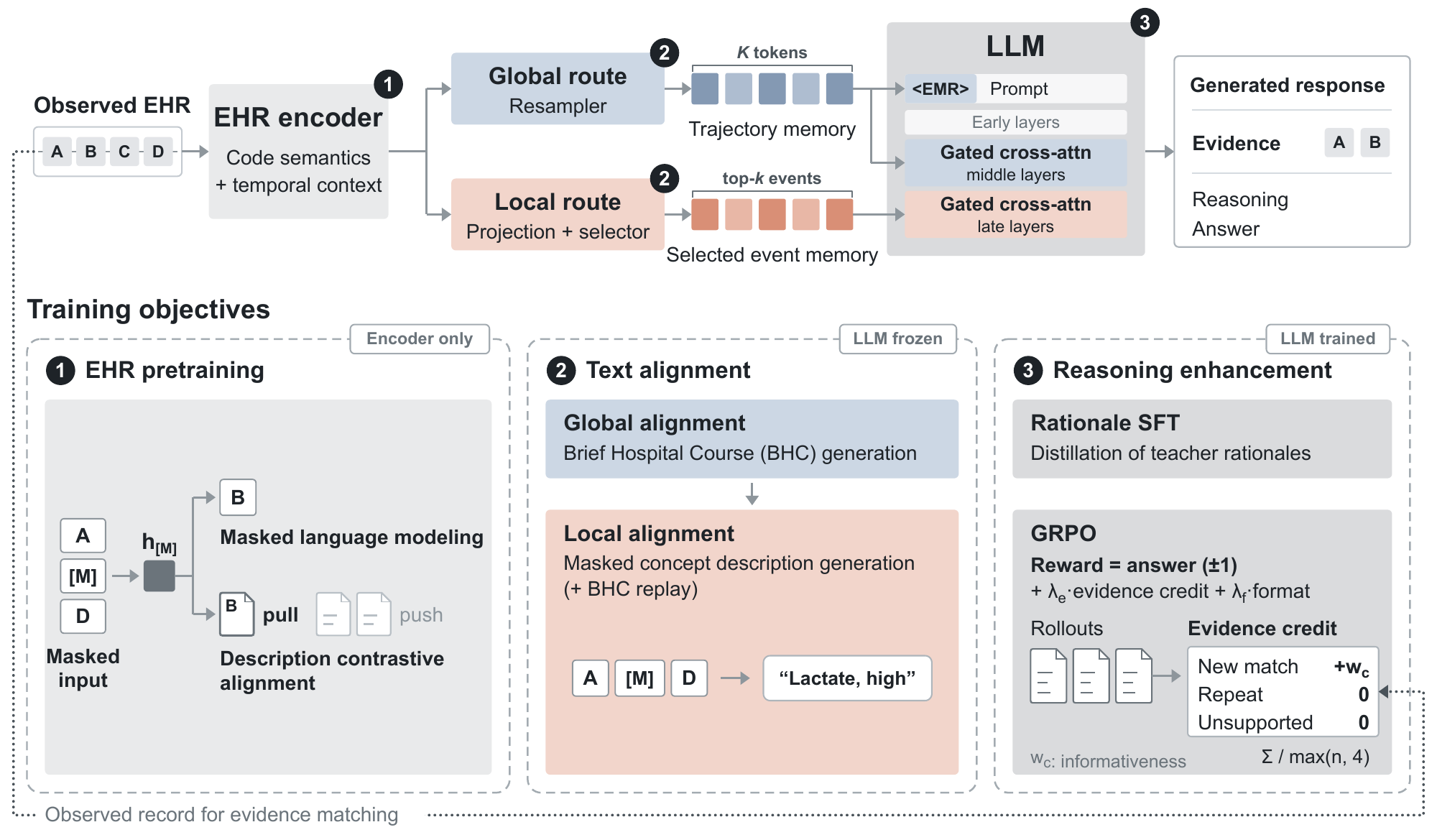}
    \caption{\textbf{Overview.}
    \textit{Top:} An EHR encoder contextualizes the observed events. The global
    route compresses them into a few tokens that fill the \texttt{<EMR>} slot and
    are re-read by gated cross-attention in the middle layers; the local route
    supplies the top-$k$ events through gated cross-attention in the late layers.
    The LLM generates cited evidence, a rationale, and an answer.
    \textit{Bottom:} \textcircled{\scriptsize 1} encoder pretraining;
    \textcircled{\scriptsize 2} trajectory (global) then concept (local)
    alignment; \textcircled{\scriptsize 3} SFT on teacher rationales followed by
    GRPO, whose evidence credit is scored against the patient's record (dotted
    line).}
    \label{fig:overview}
\end{figure}

GLoC-EHR connects an EHR encoder to a language model through two complementary
routes: a global route for trajectory-level context and a local route for
individual event-level evidence. After aligning both routes with clinical text,
we train the model for evidence-grounded reasoning through supervised
fine-tuning (SFT) on rationales followed by GRPO with answer-correctness and
evidence-support rewards.
Figure~\ref{fig:overview} summarizes the architecture and training procedure.

\paragraph{Problem setup.}
\label{sec:problem}
Given a clinical task instruction $q$, let $\mathbf{x}=(e_1,\ldots,e_T)$ be the
events retained within its observation window. Each event consists of a medical
concept $c_t$ and associated temporal and event-type attributes. Given
$(\mathbf{x},q)$, our goal is to generate a response $o=(r,a)$, where $r$
contains cited findings and a rationale grounded in $\mathbf{x}$, and $a$ is
the final task answer.

\subsection{Description-informed EHR encoder}
\label{sec:ehr_encoder}

A bidirectional EHR encoder produces contextualized event states
$\mathbf{H}=f_{\mathrm{EHR}}(\mathbf{x})\in\mathbb{R}^{T\times d_E}$.
Building on masked language modeling (MLM) \citep{devlin2019bert} pretraining
for structured EHRs \citep{li2020behrt,rasmy2021medbert,pang2021cehrbert}, it
learns these states
from the concept, age, date, and type of each event
(Appendix~\ref{app:encoder}). Prior
studies show that incorporating textual concept descriptions can improve
transfer across institutions and coding systems
\citep{kim2026medrep,kirchler2026grasp}. We therefore incorporate description
embeddings into the EHR encoder, using the backbone LLM to encode each
concept's description rather than introducing a separate text encoder.

Because the backbone LLM is not trained for text embedding, we first assessed
how well its representations distinguish concept descriptions
(Appendix~\ref{app:desc_embedding}) and found strong anisotropy in both
mean-pooled input-token embeddings and final-layer hidden states
\citep{ethayarajh2019contextual}. To address this, we applied centering,
whitening, and $\ell_2$ normalization \citep{su2021whitening,huang2021whiteningbert}
to the LLM's mean-pooled final-layer hidden states. We use the resulting near-isotropic vectors, which showed the clearest
concept structure among the evaluated representations, as frozen description
embeddings $\mathbf{z}_c$. The final event input
embedding combines the projected description embedding with a learned code-ID
embedding as a scaled residual, together with temporal and event-type
embeddings.

The encoder is pretrained with masked-concept prediction and a contrastive
objective that aligns masked event states with their target concept
descriptions. We apply code-ID dropout to discourage shortcut learning through
exclusive reliance on learned code-ID embeddings and encourage the encoder to
use description embeddings. At prediction targets, both the code-ID and
description inputs are masked to prevent target-description leakage.
The pretrained encoder is then frozen during interface
alignment and downstream adaptation. Description processing, embedding
equations, pretraining objectives, and sampling settings are specified in
Appendix~\ref{app:encoder}.

\subsection{Global and local EHR memories}
\label{sec:dual_interface}

Given the event states $\mathbf{H}$, we define two complementary routes for
injecting EHR information into the LLM. First, the global route summarizes the
overall context of the observed trajectory; second, the local route retains
individual representations of selected events as event-level evidence.

\paragraph{Global memory.}
Following latent-query resampling
\citep{jaegle2021perceiver,alayrac2022flamingo,li2023blip2}, a learned resampler uses $K_g$ queries $\mathbf{Q}$ to attend to the event
sequence and produce a fixed-size memory:
\begin{equation}
    \mathbf{M}^{\mathrm{glob}}
      =f_{\mathrm{res}}(\mathbf{Q},\mathbf{H})
      \in\mathbb{R}^{K_g\times d_L},
    \label{eq:global_memory}
\end{equation}
where $d_L$ is the LLM's hidden dimension. These global memory tokens are
inserted into the prompt as continuous embeddings and also serve as the memory
for global cross-attention. The resampler therefore provides a compact interface
whose size is independent of the number of retained events.

\paragraph{Local memory.}
Evidence-grounded reasoning requires linking generated findings to specific
recorded concepts and their descriptions. The local route therefore maps each
event state to the LLM's hidden dimension with a projection $p_{\ell}$ and ranks
it with a selector $s$:
\begin{equation}
    \mathbf{c}_t=p_{\ell}(\mathbf{H}_t),\qquad
    \mathcal{I}=\operatorname{TopK}_{K_\ell}\{s(\mathbf{c}_t):t\in\mathcal{V}\},\qquad
    \mathbf{M}^{\mathrm{loc}}=(\mathbf{c}_t)_{t\in\mathcal{I}},
    \label{eq:local_memory}
\end{equation}
where $\mathcal{V}$ is the set of eligible event positions and selection retains
up to $K_{\ell}$ events. Local vectors are accessed only through
cross-attention. The selector does
not see the task instruction, so its ranking is shared across tasks. It is first learned during concept alignment
(Section~\ref{sec:alignment}) and then adapted once to rationale generation
(Section~\ref{sec:downstream_adaptation}); it remains fixed for direct
prediction and GRPO.

\paragraph{Gated integration.}
The language model attends to each memory through separate residual modules
\citep{alayrac2022flamingo}. For route
$r\in\{\mathrm{glob},\mathrm{loc}\}$, a module after decoder block $l$ updates
the block's output hidden states $\mathbf{Z}^{(l)}$ as
\begin{equation}
    \mathbf{Z}^{(l)}\leftarrow\mathbf{Z}^{(l)}
      +\sigma(\gamma_l^r)\operatorname{XAttn}_l^r
        \bigl(\mathbf{Z}^{(l)},\mathbf{M}^r\bigr),
    \label{eq:gated_xattn}
\end{equation}
where $\operatorname{XAttn}_l^r$ is cross-attention from $\mathbf{Z}^{(l)}$ to
the memory $\mathbf{M}^r$, $\sigma$ is the logistic sigmoid, and the learned
scalar gate $\gamma_l^r$ controls the contribution of that route. Motivated by
layer-dependent cross-modal information flow \citep{zhang2025crossmodalflow},
we place global modules before local modules, making trajectory context
available before the final event-level injections. Memory budgets, projections, layer
placements, and attention masks are given in Appendix~\ref{app:interface}.

\subsection{Text alignment of the two routes}
\label{sec:alignment}

We align the global--local interface with clinical text at both the trajectory
and concept levels. For trajectory-level alignment, we pair the
structured event sequence of a hospital admission with the Brief Hospital
Course (BHC) section of its discharge summary, treating them as structured and
free-text views of the same admission. For concept-level alignment, we pair
individual medical concepts with their textual descriptions. We train the
global route first and the local route second, keeping the EHR encoder and
base language model frozen throughout alignment.

\paragraph{Trajectory alignment.}
The global route is trained to generate the paired BHC text from the structured
admission trajectory. Response-only next-token
prediction updates the resampler and global cross-attention modules, with the
local route disabled. BHC text serves as supervision for communicating
trajectory-level information; downstream inputs consist of the structured
observation window and the task instruction.

\paragraph{Concept alignment.}
We then mask clinical concepts and train the local route to recover their
textual descriptions, interleaving these examples with BHC replay. The selector
follows three phases. In \emph{forced selection}, masked event positions are
placed in local memory, and direct supervision trains the selector to identify
these positions as initial selection targets. In \emph{soft selection}, masked
positions are excluded from the candidate set, and the description-generation
loss trains the selector through differentiable memory weights to prioritize
unmasked events that help reconstruct the target descriptions. Finally,
\emph{hard selection} retains the top-$k$ events, matching the bounded local
memory used downstream. Masked positions remain excluded, and the projection
and selector are frozen while local cross-attention continues to adapt.
The full schedule, auxiliary selector losses, and trainable
components are detailed in Appendix~\ref{app:alignment}.

\subsection{Evidence-grounded reasoning with GRPO}
\label{sec:downstream_adaptation}

Supervised warm-up precedes reinforcement learning in general and medical
reasoning models \citep{guo2025deepseekr1,chen2024huatuogpto1} and, for EHRs,
establishes task competence and preserves structured clinical reasoning
\citep{lin2025ehrmind}. We therefore first perform SFT on teacher responses
containing evidence, reasoning, and an answer. The teacher receives
the observed record and gold outcome, so the targets are label-conditioned
rationales. Conditioned only on the structured EHR representations and task
instruction, the student is trained to generate the corresponding
evidence-cited rationale and task answer. SFT also adapts the local route,
because events useful for recovering masked concept descriptions need not be
those that support a rationale. As in concept alignment, the selector and local
projection are trained under soft selection, now with the rationale loss, and
then frozen under hard top-$K_\ell$ selection, which matches the memory used in
GRPO and evaluation. Teacher configuration and tuning details appear in
Appendix~\ref{app:downstream}.

\paragraph{Reward design.}
Answer correctness alone does not distinguish a supported rationale from one
that cites findings absent from the patient record. We therefore combine an
outcome signal with an explicit evidence-support term:
\begin{equation}
    R(o)=r_{\mathrm{ans}}
      +\lambda_{\mathrm{evid}}r_{\mathrm{evid}}
      +\lambda_{\mathrm{fmt}}r_{\mathrm{fmt}}
      -r_{\mathrm{len}}.
    \label{eq:reward}
\end{equation}
The answer term rewards a correct outcome, the format term encourages a
parseable response, and the length term discourages excessive generation.
Invalid answers and statements of the outcome itself (e.g., that the patient
died) receive a fixed failure reward below any valid total.

The evidence term scores the bullets listed under the evidence heading of the
response; the reasoning prose is not scored. A lexical matcher counts a bullet
as \emph{supported} if its content words match the typed description of an
event in the patient's input window, by containment or an IDF-weighted overlap
of at least $0.55$, with the same direction, negation, and family-history
attribution (Appendix~\ref{app:reward}). Matching covers every event in the
input window, not only the $K_\ell$ events in local memory, so the reward itself
does not favor the local route; descriptions and concept IDs are used only for
scoring and are never shown to the policy.

Let $f(c)$ be the fraction of training records containing concept $c$. The
$i$-th of $m$ bullets receives credit $\xi_i=1-f(c)$ if it is the first
supported citation of $c$ in the response. Unsupported bullets, repeated
citations, and non-clinical metadata such as age or sex receive zero credit.
The evidence reward is
\begin{equation}
    S=\frac{1}{\max(m,m_0)}\sum_{i=1}^{m}\xi_i,\qquad
    r_{\mathrm{evid}}=2S-1
    \quad (m>0),
    \label{eq:evidence_reward}
\end{equation}
with $m_0=4$ and $r_{\mathrm{evid}}=-1$ when no evidence is listed.
Informativeness weighting keeps ubiquitous findings, such as a routine
laboratory panel, from earning full credit; repeated and unsupported bullets
lower the score through the denominator, and the floor $m_0$ removes the
gain from listing fewer than $m_0$ findings. Because the term does not reward long
lists, we report support precision together with the number of distinct
supported findings.

\paragraph{Group relative policy optimization.}
Starting from supervised initialization, we sample a group of responses to each
patient--task prompt and standardize their rewards within the group
\citep{shao2024deepseekmath}. Responses with higher relative reward are
reinforced, and a reference-policy KL term regularizes the update. Two
auxiliary losses complement the reward. The answer loss is the cross-entropy of
the gold answer at the answer position; it supplies a correctness signal even when all
responses in a group give the same answer, so that the answer reward does not
vary within the group. The
evidence-sufficiency loss is the cross-entropy of the gold answer given only the
task prompt and the response's evidence bullets, with the reasoning prose
removed and both EHR memories switched off, as in select-then-predict
rationales \citep{lei2016rationalizing,jain2020fresh}; it asks whether the cited
findings are informative about the outcome, and it is a loss rather than a
reward because the reward form favored shorter evidence lists. The encoder and memory constructors remain frozen; the language
model and cross-attention interfaces are updated. Reward coefficients, matcher
rules, sampling settings, and the precise policy-update implementation are
specified in Appendix~\ref{app:reward}. To examine whether cited local
representations affect the model's answer, we also define a matched
regeneration intervention alongside support and coverage measurements
(Appendix~\ref{app:intervention}).

\section{Experiments}
\label{sec:experiments}

\subsection{Data and tasks}
\label{sec:data}

Our main dataset is MIMIC-IV v2.2 \citep{johnson2023mimiciv}, converted to the
OMOP Common Data Model (CDM) \citep{hripcsak2015ohdsi}, with BHC targets from MIMIC-IV-Ext-BHC
\citep{aali2025bhc}. It comprises the 128,350 patients with at least one
admission that has a BHC. Patients are split 7:1:2 into training, validation,
and test sets; every stage, from alignment to the downstream tasks, uses this
split and trains only on training patients. We predict
three common binary outcomes \citep{rajkomar2018scalable}: in-hospital mortality and long length of stay (more than
7 days) from the first 24 hours of an inpatient admission, and 30-day
readmission from the full index admission. EHRSHOT \citep{wornow2023ehrshot}
serves as an external validation cohort without retraining.
Appendix~\ref{app:data} gives the cohort flow, task definitions, and sample
sizes.

\subsection{Outcome-prediction comparison}
\label{sec:outcome_comparison}

Table~\ref{tab:outcome_prediction} compares binary outcome prediction on the
three internal tasks. Prediction serves as an outcome-level control for the proposed EHR--language interface.

\paragraph{Baselines.}
\emph{Zero-shot LLMs} (Qwen3-1.7B, the backbone of GLoC-EHR, and
Llama-3.3-70B-Instruct \citep{grattafiori2024llama3}) read the records serialized as text
\citep{hegselmann2026ehrencoders} and answer with the ClinicalBench instruction
\citep{chen2024clinicalbench}, directly or after chain-of-thought (CoT)
reasoning \citep{wei2022cot}. \emph{Task-specific} models are trained per task: count-based
gradient-boosted trees \citep{ke2017lightgbm,chen2016xgboost}, RETAIN
\citep{choi2016retain}, classifier heads on frozen text embeddings
\citep{zhang2025qwen3embedding} of the serialized record and/or frozen EHR-BERT vectors
(adapted from \citealp{hegselmann2026ehrencoders}), fine-tuned EHR-BERT, and
GLoC-EHR answering directly (ST). \emph{Multi-task} models are trained once on
all three tasks: EHR-BERT (MT) uses task-specific heads on a shared encoder,
whereas GLoC-EHR (MT) and GLoC-EHR (Reasoning) receive the task only through the
instruction, and the latter writes evidence-cited reasoning before its answer.
Table~\ref{tab:outcome_prediction} shows LightGBM and the frozen-vector
baseline that uses all three feature sets; the others and per-task AUPRC are in
Appendix~\ref{app:prediction_scores}.

\paragraph{Protocol.}
Supervised models are selected by validation AUROC (macro AUROC for multi-task
models) and retrained with seeds 42--46; we report the test mean and SD,
averaging the three tasks within each seed for macro scores. Zero-shot LLMs are
evaluated once. GLoC-EHR (Reasoning) is a single GRPO run, for which we report
the mean and SD over five sampled chains per case, reflecting decoding rather
than training variability. Direct GLoC-EHR variants score only the answer
tokens, renormalized over Yes/No (Appendix~\ref{app:downstream}).

\paragraph{Results.}
Zero-shot LLMs fall far below every trained model, consistent with
ClinicalBench \citep{chen2024clinicalbench}, where LLMs did not outperform
traditional machine-learning models on mortality, length-of-stay, and
readmission prediction; CoT prompting lowers their AUROC further.
Among task-specific models, GLoC-EHR (ST) attains the highest macro AUROC and
AUPRC (0.879 and 0.549), mainly through readmission, and stays close to
fine-tuned EHR-BERT (ST) on mortality and long length of stay. Sharing one model
across the three tasks lowers GLoC-EHR's scores slightly, yet GLoC-EHR (MT)
still exceeds every baseline in macro AUROC, including EHR-BERT (ST), although
its macro AUPRC falls below those of the EHR-BERT models and several tree-based
baselines (Table~\ref{tab:outcome_prediction_full_auprc}). Two single-run
controls support the interface (Appendix~\ref{app:controls}): training it from
random initialization without alignment lowers AUROC on all three tasks, and
LoRA \citep{hu2022lora} fine-tuning of the backbone on the serialized record falls below GLoC-EHR
(MT). GLoC-EHR (Reasoning) trails GLoC-EHR (MT) by 0.008 in macro AUROC,
staying close to EHR-BERT (ST), and by 0.042 in macro AUPRC, with the AUPRC gap
concentrated on mortality and long length of stay.
Section~\ref{sec:evidence} separately evaluates the support and
informativeness of the generated evidence.

\begin{table}[!htbp]
    \centering
    \caption{Internal outcome prediction on MIMIC-IV (test): mean $\pm$ SD over five seeds, or over five sampled chains of one run for GLoC-EHR (Reasoning). Best in bold, second underlined. $C$: code counts; $T_{4096}$: text embedding; $E_{\mathrm{CLS}}$: frozen EHR-BERT vector; / LGBM: LightGBM head.}
    \label{tab:outcome_prediction}
    \begingroup
    \footnotesize
    \setlength{\tabcolsep}{1pt}
    \renewcommand{\arraystretch}{1.1}
    \newcommand{\predscore}[2]{\ensuremath{#1{\scriptstyle\,\pm\,#2}}}
    \begin{tabular*}{\linewidth}{@{\extracolsep{\fill}}lccccc@{}}
        \toprule
        & \multicolumn{3}{c}{AUROC $\uparrow$} & \multicolumn{2}{c}{Macro (3 tasks) $\uparrow$} \\
        \cmidrule(lr){2-4}\cmidrule(l){5-6}
        Method & Mortality & Long LOS & Readmission & AUROC & AUPRC \\
        & {\scriptsize (24\,h)} & {\scriptsize (24\,h)} & {\scriptsize (full)} & & \\
        \midrule
        \multicolumn{6}{@{}l}{\textit{Zero-shot LLMs on serialized records}} \\
        Qwen3-1.7B (direct) & \ensuremath{0.8140} & \ensuremath{0.6062} & \ensuremath{0.5183} & \ensuremath{0.6461} & \ensuremath{0.2240} \\
        Qwen3-1.7B (CoT) & \ensuremath{0.5488} & \ensuremath{0.5363} & \ensuremath{0.5052} & \ensuremath{0.5301} & \ensuremath{0.1631} \\
        Llama-3.3-70B (direct) & \ensuremath{0.8268} & \ensuremath{0.6391} & \ensuremath{0.5269} & \ensuremath{0.6643} & \ensuremath{0.2065} \\
        Llama-3.3-70B (CoT) & \ensuremath{0.6780} & \ensuremath{0.5193} & \ensuremath{0.5034} & \ensuremath{0.5669} & \ensuremath{0.1644} \\
        \addlinespace[2pt]
        \multicolumn{6}{@{}l}{\textit{Task-specific models}} \\
        LightGBM (counts) & \predscore{0.9527}{0.0013} & \predscore{0.8624}{0.0011} & \predscore{0.7758}{0.0028} & \predscore{0.8637}{0.0013} & \predscore{0.5299}{0.0102} \\
        $C{+}T_{4096}{+}E_{\mathrm{CLS}}$ / LGBM & \predscore{0.9449}{0.0027} & \predscore{0.8589}{0.0013} & \predscore{0.7625}{0.0068} & \predscore{0.8554}{0.0017} & \predscore{0.4940}{0.0042} \\
        EHR-BERT (ST) & \predscore{\mathbf{0.9607}}{0.0018} & \predscore{\underline{0.8701}}{0.0008} & \predscore{0.7817}{0.0057} & \predscore{0.8708}{0.0020} & \predscore{\underline{0.5380}}{0.0034} \\
        GLoC-EHR (ST) & \predscore{0.9561}{0.0017} & \predscore{\mathbf{0.8740}}{0.0018} & \predscore{\mathbf{0.8075}}{0.0015} & \predscore{\mathbf{0.8792}}{0.0008} & \predscore{\mathbf{0.5494}}{0.0066} \\
        \addlinespace[2pt]
        \multicolumn{6}{@{}l}{\textit{Multi-task models}} \\
        EHR-BERT (MT) & \predscore{\underline{0.9575}}{0.0017} & \predscore{0.8672}{0.0019} & \predscore{0.7865}{0.0026} & \predscore{0.8704}{0.0013} & \predscore{0.5312}{0.0062} \\
        GLoC-EHR (MT) & \predscore{0.9545}{0.0013} & \predscore{0.8671}{0.0030} & \predscore{0.7976}{0.0052} & \predscore{\underline{0.8731}}{0.0030} & \predscore{0.5236}{0.0088} \\
        GLoC-EHR (Reasoning) & \predscore{0.9459}{0.0010} & \predscore{0.8469}{0.0004} & \predscore{\underline{0.8042}}{0.0018} & \predscore{0.8656}{0.0008} & \predscore{0.4819}{0.0045} \\
        \bottomrule
    \end{tabular*}
    \endgroup
\end{table}

\subsection{Evidence grounding}
\label{sec:evidence}

\paragraph{Metrics.}
We score the evidence bullets of every sampled chain with the matcher used by
the reward (Section~\ref{sec:downstream_adaptation}; Appendix~\ref{app:reward}).
\emph{Support precision} is the informativeness-weighted fraction of bullets
that cite a distinct supported finding, $\frac{1}{m}\sum_i\xi_i$. The
\emph{unsupported rate} is the fraction of bullets that match no event in the
input window, and \emph{distinct supported findings} counts the concepts
credited in a response. \emph{Bullets-only AUROC} measures how informative the cited
evidence is to the model that wrote it: the model answers from the task prompt
and its evidence bullets alone, with both EHR memories switched
off. Test-set values average the five chains of each case. Over GRPO training,
the unsupported rate falls for GLoC-EHR and the global-route-only variant but
stays near its SFT level without the evidence terms
(Appendix~\ref{app:training_dynamics}).

\begin{figure}[t]
    \centering
    \includegraphics[width=\linewidth]{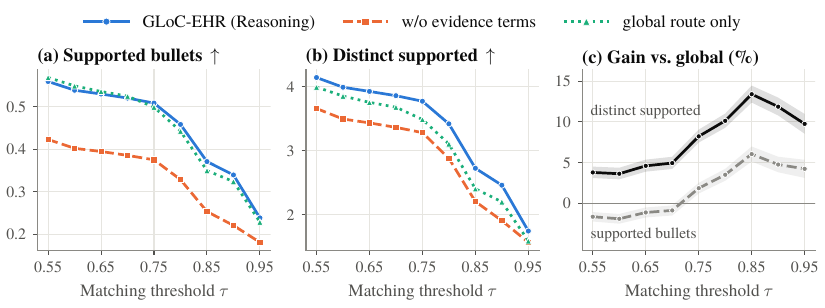}
    \caption{Evidence of the final checkpoints on MIMIC-IV test chains (one per case) as matching becomes
    stricter: every match must reach an IDF-weighted overlap of at least $\tau$ ($\tau{=}0.55$ is the reward's
    matcher). (a) Share of supported bullets, (b) distinct supported findings per response, and (c) relative gain
    of GLoC-EHR (Reasoning) over the global-route-only variant, with 95\% paired case-bootstrap intervals.}
    \label{fig:evidence_curves}
\end{figure}

\paragraph{Held-out evidence.}
On the test chains (Table~\ref{tab:reasoning_ablation}; brackets: 95\% paired
case-bootstrap intervals), GRPO lowers the unsupported rate from 0.58 after
rationale SFT to 0.44 while writing fewer bullets and citing more distinct
supported findings. Without the evidence terms, the unsupported rate stays at
the SFT level, $0.137$ [$0.133$, $0.140$] above that of GLoC-EHR, and fewer
distinct supported findings are cited. The global-route-only variant writes
fewer bullets with a slightly higher support precision but cites $0.15$
[$0.12$, $0.18$] fewer distinct supported findings per response.
Figure~\ref{fig:evidence_curves} re-scores these chains under stricter matching,
which requires closer lexical agreement between a bullet and the event it cites.
From $\tau=0.75$ onward GLoC-EHR has a higher share of supported bullets than
the global-route-only variant, and at $\tau=0.85$ it cites 13.4\% [12.4, 14.5] more
distinct supported findings. The local route thus yields citations that match their events more closely. Zero-shot LLMs that read the serialized
record can reach high support by quoting it but answer near chance
(Appendix~\ref{app:reasoning_snapshot}).

\paragraph{Cited-slot masking.}
To test whether the local route supplies the cited evidence, we mask the
local-memory slots of the concepts that a greedy chain cites, regenerate the
chain, and compare with masking as many uncited slots of similar
informativeness (2{,}062 test cases whose chains cite local-memory concepts;
Appendix~\ref{app:intervention}). The regenerated rationale drops the masked
concepts more often under the cited mask (persistence $-0.017$ [$-0.028$,
$-0.007$]), but the answer changes about as much as under the control mask (macro
AUROC $-0.003$ [$-0.009$, $0.002$]; Table~\ref{tab:mediation}). The local route
thus shapes which findings are cited, with no detectable effect on the answer,
consistent with the similar AUROC of the global-route-only variant (Section~\ref{sec:ablations}).

\subsection{Ablations}
\label{sec:ablations}

Table~\ref{tab:reasoning_ablation} compares GLoC-EHR (Reasoning) with variants
that remove the evidence terms (the evidence reward and the evidence-sufficiency
loss), use the global route only, or stop after rationale SFT. Stopping after
rationale SFT lowers macro AUROC from 0.866 to 0.658, and the SFT model answers
almost deterministically (99.8\% of its answer probabilities are below 0.02 or
above 0.98, vs.~31\% after GRPO). Neither removing the evidence terms nor
dropping the local route detectably changes macro AUROC (paired bootstrap intervals include
zero), but the bullets alone predict the outcome slightly better with the global
route only and worse without the evidence terms, a variant that, like rationale
SFT, is never trained to answer from its bullets.

\begin{table}[t]
    \centering
    \caption{Ablations of GLoC-EHR (Reasoning) on the MIMIC-IV test set. The first row is the mean $\pm$ SD over
    five sampled chains; the other rows score one chain per case. Evidence is matched with the reward's matcher
    ($\tau{=}0.55$) and, for distinct supported findings, also at $\tau{=}0.85$ (Figure~\ref{fig:evidence_curves}).
    Bullets-only AUROC: answers from each model's own bullets with the EHR memories off.}
    \label{tab:reasoning_ablation}
    \begingroup
    \footnotesize
    \setlength{\tabcolsep}{2pt}
    \newcommand{\predscore}[2]{\ensuremath{#1{\scriptstyle\,\pm\,#2}}}
    \begin{tabular*}{\linewidth}{@{\extracolsep{\fill}}lcccccc@{}}
        \toprule
        & Macro & Macro & Unsupported & \multicolumn{2}{c}{Distinct supported $\uparrow$} & Bullets-only \\
        \cmidrule(lr){5-6}
        Variant & AUROC $\uparrow$ & AUPRC $\uparrow$ & rate $\downarrow$ & $\tau{=}0.55$ & $\tau{=}0.85$ & AUROC $\uparrow$ \\
        \midrule
        GLoC-EHR (Reasoning) & \predscore{0.8656}{0.0008} & \predscore{0.4819}{0.0045} & \predscore{0.440}{0.001} & \predscore{4.16}{0.01} & \predscore{2.73}{0.01} & \predscore{0.758}{0.005} \\
        \quad one chain & 0.8661 & 0.4810 & 0.440 & 4.15 & 2.73 & 0.755 \\
        \quad w/o evidence terms & 0.8665 & 0.4896 & 0.577 & 3.66 & 2.21 & 0.707 \\
        \quad global route only & 0.8641 & 0.4783 & 0.431 & 4.00 & 2.41 & 0.767 \\
        \quad rationale SFT only & 0.6578 & 0.2013 & 0.582 & 3.92 & 2.34 & 0.690 \\
        \bottomrule
    \end{tabular*}
    \endgroup
\end{table}

\subsection{Transferability}
\label{sec:external}

Table~\ref{tab:external_main} applies the internally selected checkpoints to
EHRSHOT without retraining, after removing events with codes outside the
training vocabulary; an EHR-BERT pretrained on code IDs alone isolates the
contribution of description embeddings. All three main models lose about 0.1
macro AUROC relative to their internal test scores, so GLoC-EHR (MT) remains on
par with EHR-BERT (MT) and the reasoning variant stays close to both, while
EHR-BERT (MT) keeps the highest macro AUPRC. Without description embeddings,
EHR-BERT falls further on readmission, the hardest task for every model, which
accounts for nearly all of its lower macro AUROC. Asked, without further
training, whether a patient's next serum sodium or platelet count will be low,
GLoC-EHR (Reasoning) reaches AUROC 0.657 and 0.629 on held-out MIMIC-IV
admissions, compared with 0.556 and 0.525 for zero-shot Qwen3-1.7B reading the
serialized record (Appendix~\ref{app:transfer}).

\begin{table}[t]
    \centering
    \caption{External validation on EHRSHOT without retraining, with out-of-vocabulary events removed: mean $\pm$
    SD over five seeds. GLoC-EHR (Reasoning): one sampled chain per case (internal five-chain SD 0.0008), excluding the
    2.6\% without an answer (0.767 macro AUROC if scored as 0.5). Code IDs only: EHR-BERT pretrained without
    description embeddings. Best in bold, second underlined.}
    \label{tab:external_main}
    \begingroup
    \footnotesize
    \setlength{\tabcolsep}{1pt}
    \newcommand{\predscore}[2]{\ensuremath{#1{\scriptstyle\,\pm\,#2}}}
    \begin{tabular*}{\linewidth}{@{\extracolsep{\fill}}lccccc@{}}
        \toprule
        & \multicolumn{3}{c}{AUROC $\uparrow$} & \multicolumn{2}{c}{Macro $\uparrow$} \\
        \cmidrule(lr){2-4}\cmidrule(l){5-6}
        Method & Mortality & Long LOS & Readmission & AUROC & AUPRC \\
        \midrule
        EHR-BERT (MT) & \predscore{\mathbf{0.9311}}{0.0055} & \predscore{0.7531}{0.0056} & \predscore{\underline{0.6314}}{0.0153} & \predscore{\underline{0.7719}}{0.0041} & \predscore{\mathbf{0.4414}}{0.0040} \\
        \quad code IDs only & \predscore{\underline{0.9303}}{0.0037} & \predscore{0.7525}{0.0063} & \predscore{0.5568}{0.0328} & \predscore{0.7465}{0.0130} & \predscore{0.4167}{0.0148} \\
        GLoC-EHR (MT) & \predscore{0.9287}{0.0023} & \predscore{\underline{0.7584}}{0.0043} & \predscore{\mathbf{0.6396}}{0.0131} & \predscore{\mathbf{0.7756}}{0.0039} & \predscore{\underline{0.4331}}{0.0060} \\
        GLoC-EHR (Reasoning) & $0.9244$ & $\mathbf{0.7585}$ & $0.6288$ & $0.7706$ & $0.4222$ \\
        \bottomrule
    \end{tabular*}
    \endgroup
\end{table}

\section{Conclusion}
\label{sec:conclusion}

GLoC-EHR gives a language model a fixed-size global memory of an EHR trajectory
and a bounded local memory of selected events, aligns both with clinical text,
and trains evidence-cited reasoning with rewards for correct answers and
record-supported evidence. Answering directly, it matches or exceeds fine-tuned
EHR encoders in macro AUROC on three MIMIC-IV tasks, far above zero-shot LLMs.
Citing evidence first costs little discrimination after GRPO (0.008 macro AUROC
below the direct multi-task model, with a larger AUPRC gap), whereas rationale
SFT alone reaches only 0.658. At a similar macro AUROC, the evidence terms lower
the share of unsupported evidence from 0.58 to 0.44, and the local route adds
distinct supported findings and shapes which findings are cited, with no detectable effect on the
answer. Without retraining, GLoC-EHR transfers to EHRSHOT on par with EHR-BERT
and answers unseen laboratory questions better than its zero-shot backbone.

\paragraph{Limitations and future work.}
Our evidence matching is lexical: it checks that each cited finding appears in
the patient's record, but not whether the finding is clinically correct or
whether the model relied on it to answer. Future work includes semantic matching
of paraphrased findings, for both the reward and the evaluation, and clinician
review of the cited evidence. GLoC-EHR (Reasoning) comes from a single training
run on data from one health system; additional runs and cohorts would further
test the stability of these effects. We also plan to recalibrate its
probabilities for each new cohort: on EHRSHOT, the mean predicted probability of
a long stay was 0.57 against a prevalence of 0.31.

\subsection*{AI use statement}
In this work, we used generative AI tools to assist with translation and to
provide feedback on the design of the methodology and experiments. Apart from the teacher rationales used as SFT targets, which
are part of our method (Section~\ref{sec:downstream_adaptation}), we have not
used generative AI tools to generate synthetic data sets, and tasks involving
mathematical claims, proofs, or qualitative and thematic data analysis are not
applicable to this work. Additionally, we used generative AI tools to draft
parts of the paper and edit it for readability, to create and edit analysis and
plotting code and figures, and to search for and identify relevant literature.
We have reviewed all AI-assisted work: the authors wrote the main content of the
paper and revised all AI-drafted text, read and verified every cited work, and
checked all AI-assisted code and results. We take responsibility for the final
content of this work, including text, claims or artifacts produced with the aid
of generative AI.

\subsection*{Ethics statement}
This work uses only de-identified EHR data. MIMIC-IV and MIMIC-IV-Ext-BHC were
accessed through PhysioNet under its credentialed data use agreement, and
EHRSHOT under its research data use agreement; no new data were collected, and
no attempt was made to re-identify patients. All language models, including the
teacher and the zero-shot baselines, were run on the authors' servers, so no
patient data were sent to external services. GLoC-EHR is a research model and
is not intended for clinical decision-making. Its evidence is checked for
lexical support in the record, not for clinical correctness; its probabilities
are not calibrated on the external cohort
\citep{guo2017calibration,vancalster2019calibration}; and it was trained on data
from a single health system, whose patient population and documentation
practices it may reflect. Because clinical models can degrade under dataset and
temporal shift \citep{finlayson2021datasetshift,guo2023robustness}, clinical use
would require prospective validation, recalibration, and evaluation across
patient subgroups.

\subsection*{Reproducibility statement}
\paragraph{Data.}
MIMIC-IV v2.2, MIMIC-IV-Ext-BHC, and EHRSHOT are available to credentialed
researchers under their data use agreements. Appendix~\ref{app:data} describes
the cohort construction, task definitions, patient-level splits, and sample
sizes.

\paragraph{Method.}
Section~\ref{sec:method} describes the architecture and training stages, and
Appendix~\ref{app:implementation} specifies the encoder, the interface, the
alignment schedule, rationale SFT, the GRPO settings, the reward coefficients,
and the full evidence matcher. GRPO training of GLoC-EHR (Reasoning) took about
55 hours on four NVIDIA H200 GPUs.

\paragraph{Evaluation.}
Section~\ref{sec:experiments} and Appendix~\ref{app:prediction_scores} specify
the metrics, seeds, sampling settings, bootstrap procedures, and the handling
of chains without an answer. Main results are reported as means and standard
deviations over five seeds or five sampled chains.

\paragraph{Code and model weights.}
We will release the code upon acceptance. Because the models were trained on
credentialed data, we will release the model weights upon acceptance through
PhysioNet under the same credentialed-access terms as MIMIC-IV.

\bibliography{iclr2027_conference}
\bibliographystyle{iclr2027_conference}

\appendix
\section{Implementation and evaluation details}
\label{app:implementation}

This appendix records the configuration used in the current implementation.
The architectural design and learning objectives are described in
Section~\ref{sec:method}; numerical settings and evaluation procedures are
collected here for reproducibility.

\subsection{Structured input and encoder pretraining}
\label{app:encoder}

\paragraph{Input representation.}
Each example contains at most $T_{\max}=2048$ structured events after applying
its task-specific observation cutoff and truncation rule. Each event contains
a medical concept together with parallel age, date, and event-type attributes.
The date field is the rank of the event date within the patient's record,
rather than an absolute calendar date. The observed sequence is therefore not
the patient's unrestricted longitudinal history. The vocabulary contains
20,561 tokens constructed from the OMOP domains represented in our MIMIC-IV
cohort \citep{johnson2023mimiciv}. Measurements carry qualitative or quantile value suffixes, and drug
concepts carry exposure-duration suffixes. Cohort construction, task
definitions, and the patient-level split shared by all training stages are
described in Appendix~\ref{app:data}.

\paragraph{Description embeddings.}
For each vocabulary concept $c$, a short typed description $d(c)$ is formed,
such as ``condition: Enteropathic arthritis'' or ``lab: Sodium [Moles/volume] in
Serum or Plasma, low''. We feed at most 128 description tokens to Qwen3-1.7B
\citep{qwen3report} and mean-pool its final hidden states to obtain a
2,048-dimensional vector $\mathbf{b}_c$. We center, whiten, and
$\ell_2$-normalize the description vectors:
\begin{equation}
    \mathbf{z}_c =
    \operatorname{norm}_2\!\left[
       (\boldsymbol{\Sigma}+\epsilon\mathbf{I})^{-1/2}
       (\mathbf{b}_c-\boldsymbol{\mu})
    \right].
    \label{eq:whitening}
\end{equation}
The mean and covariance are estimated once over the vocabulary descriptions,
which uses no patient data. Whitening follows related normalization for
sentence representations \citep{huang2021whiteningbert}; the comparison that
motivated it is reported in Appendix~\ref{app:desc_embedding}. The description
table $\{\mathbf{z}_c\}$ remains frozen during EHR pretraining.

\paragraph{Event embeddings and encoder.}
The EHR encoder is an eight-layer, 12-head bidirectional Transformer with
hidden size $d_E=768$ and no absolute-position embedding. Let $\mathbf{v}_c$
be a learned code-ID embedding, and let $\mathbf{a}_t$,
$\boldsymbol{\delta}_t$, and $\boldsymbol{\tau}_t$ denote learned age, date,
and type embeddings. The input at position $t$ is
\begin{align}
    \widetilde{\mathbf{d}}_{c_t}
        &= \operatorname{LN}_d(\mathbf{W}_d\mathbf{z}_{c_t}+\mathbf{b}_d),\\
    \mathbf{u}_t
        &= \operatorname{LN}\!\left(
        \widetilde{\mathbf{d}}_{c_t}
        +0.3\operatorname{LN}_c(\mathbf{v}_{c_t})
        +\mathbf{a}_t+\boldsymbol{\delta}_t+\boldsymbol{\tau}_t
        \right).
    \label{eq:event_embedding}
\end{align}
The description is thus the base representation and the learned code ID is a
scaled residual. At eligible non-MLM context positions, the input code ID is
replaced by \texttt{[UNK]} with probability $0.5$, while retaining the correct
description-table index. At masked positions, both the code and description
indices point to \texttt{[MASK]}, preventing target-description leakage.

\paragraph{Sampling and pretraining objective.}
We sample one view per patient per epoch: a single visit with probability
$0.6$, the most recent trajectory with probability $0.3$, or a random contiguous
visit window with probability $0.1$. Each view is capped at 2,048 events.
We mask $30\%$ of eligible clinical-concept positions and replace every selected
position with \texttt{[MASK]}, without an 80/10/10 corruption rule. Besides
masked concept prediction, each masked contextual state is contrasted against
the projected description of its true concept. For $M$ masked positions in a
batch, let $\mathbf{h}_i$ be the normalized encoder state and
$\widetilde{\mathbf{z}}_i$ the normalized projected target description. The
one-directional in-batch contrastive objective is
\begin{equation}
    \mathcal{L}_{\mathrm{con}}=-\frac{1}{M}\sum_{i=1}^{M}
    \log\frac{\exp(\mathbf{h}_i^\top\widetilde{\mathbf{z}}_i/\tau)}
    {\sum_{j=1}^{M}\exp(\mathbf{h}_i^\top\widetilde{\mathbf{z}}_j/\tau)},
    \qquad \tau=0.1.
\end{equation}
We optimize
\begin{equation}
    \mathcal{L}_{\mathrm{EHR}}
      =\mathcal{L}_{\mathrm{MLM}}+0.1\mathcal{L}_{\mathrm{con}}.
    \label{eq:ehr_pretrain}
\end{equation}
The resulting contextual sequence is
$\mathbf{H}=f_{\mathrm{EHR}}(\mathbf{x})\in\mathbb{R}^{T\times d_E}$.

\subsection{Backbone description embeddings}
\label{app:desc_embedding}

Description-based concept representations improve the transfer of EHR models
across vocabularies and institutions, using LLM-generated definitions combined
with ontology structure \citep{kim2026medrep} or frozen embeddings from a
dedicated text-embedding model \citep{kirchler2026grasp}. GLoC-EHR later maps
encoder states into Qwen3-1.7B, so we compute the description embeddings with
that backbone. The encoder and the language model then share one semantic
reference rather than relying on a third embedding space. Because Qwen3-1.7B is
a generative model rather than an embedding model, and decoder-only LLMs
typically need adaptation to serve as text encoders
\citep{behnamghader2024llm2vec}, we compared several ways of
turning a description into a vector before fixing Equation~\ref{eq:whitening}
(Table~\ref{tab:desc_embedding_geometry}).

\paragraph{Setup.}
All 20{,}561 vocabulary tokens receive a typed short description; special and
metadata tokens receive brief labels such as ``masked EHR code''. We mean-pool
either the input-embedding rows of the description tokens (\emph{input-token
mean}) or the final-layer hidden states of a forward pass over at most 128
tokens (\emph{final hidden state}). We then apply no post-processing, removal
of the first principal component \citep{mu2018allbutthetop} followed by $\ell_2$
normalization, or centering, whitening, and $\ell_2$ normalization
\citep{su2021whitening}. Isotropy is summarized by the
entropy effective rank of the covariance spectrum, the variance share of the
first principal component, and the mean pairwise cosine. Qualifier separation
uses the 6{,}952 tokens that attach a qualifier to a base concept: 1{,}757
measurement tokens with value bins (e.g., low, high, quartile) and 5{,}195 drug
tokens with short or long exposure. For each, we find its nearest neighbour by
cosine among all vocabulary tokens and record whether it is another variant of
the same concept or a different concept with the same qualifier.

\paragraph{Findings.}
Raw representations are strongly anisotropic, as observed for other pretrained
language models \citep{ethayarajh2019contextual,li2020bertflow}. Input-token means occupy an
effective rank of 295 of 2{,}048 dimensions, and final hidden states collapse
further to 70, with a mean pairwise cosine of 0.89. In the raw input space,
random qualified concepts have a cosine of 0.61, and 48\% of qualified tokens
have a different concept with the same qualifier as their nearest neighbour.
Whitening restores isotropy (effective rank 1{,}793 for input-token means and
1{,}908 for final hidden states) and drives the cosine between unrelated
concepts to zero while keeping variants of the same concept close (0.94).
Removing only the first principal component leaves the space anisotropic
(effective rank 116, random-pair cosine 0.36). Whitened final hidden states give
the clearest concept structure: 98.4\% of qualified tokens that have a sibling
variant find it as their nearest neighbour, compared with 91.1\% for whitened
and 75.3\% for raw input-token means, and the share of same-qualifier
neighbours is lowest (31.9\%). The forward pass thus composes a multi-token
description into a concept-level vector that qualifier words perturb less than
a token average does.

The same property means that whitened description embeddings mainly encode
concept identity. For ``lab: Sodium [Moles/volume] in Serum or Plasma, low'', the three nearest
neighbours are the high, very low, and unqualified variants of the same test. Value and exposure
distinctions are therefore carried chiefly by the learned code-ID residual and
the surrounding events (Equation~\ref{eq:event_embedding}), which the
contextual encoder combines during pretraining.

\begin{table}[t]
    \centering
    \caption{Geometry of Qwen3-1.7B description embeddings for the 20{,}561 vocabulary concepts (typed short
    descriptions, mean pooling, 2{,}048 dimensions). Isotropy: effective rank $r_{\mathrm{eff}}$, variance share of the first
    principal component, and mean pairwise cosine $\bar{c}$, which is zero by construction after centring. Qualified
    concepts: the 6{,}952 concepts with a value or exposure qualifier; $\bar{c}_{\mathrm{same}}$ and $\bar{c}_{\mathrm{rand}}$
    are mean cosines between variants of the same concept and between random qualified concepts; NN$_{\mathrm{c}}$ is the
    share of nearest neighbours that are a variant of the same concept, among the 68.5\% of qualified concepts that have
    one; NN$_{\mathrm{q}}$ is the share whose nearest neighbour is a different concept with the same qualifier. Without the
    type prefix, the used embeddings give an NN$_{\mathrm{c}}$ of 97.1\%, and float32 inference changes NN$_{\mathrm{c}}$ by
    less than 0.2 points.}
    \label{tab:desc_embedding_geometry}
    \begingroup
    \footnotesize
    \setlength{\tabcolsep}{3pt}
    \begin{tabular*}{\linewidth}{@{\extracolsep{\fill}}llrrrrrrr@{}}
        \toprule
        & & \multicolumn{3}{c}{Isotropy} & \multicolumn{4}{c}{Qualified concepts} \\
        \cmidrule(lr){3-5}\cmidrule(l){6-9}
        Representation & Post-processing & $r_{\mathrm{eff}}\uparrow$ & PC$_1$\,\% $\downarrow$ & $\bar{c}\downarrow$ & $\bar{c}_{\mathrm{same}}$ & $\bar{c}_{\mathrm{rand}}\downarrow$ & NN$_{\mathrm{c}}$\,\% $\uparrow$ & NN$_{\mathrm{q}}$\,\% $\downarrow$ \\
        \midrule
        Input-token mean & none & 295 & 15.3 & 0.29 & 0.97 & 0.61 & 75.3 & 48.4 \\
        Input-token mean & whiten $+\,\ell_2$ & 1{,}793 & 0.2 & 0.00 & 0.94 & 0.00 & 91.1 & 36.4 \\
        \addlinespace[2pt]
        Final hidden state & none & 70 & 21.3 & 0.89 & \textemdash & \textemdash & \textemdash & \textemdash \\
        Final hidden state & top-PC removal $+\,\ell_2$ & 116 & 23.4 & 0.00 & 0.99 & 0.36 & 87.0 & 39.5 \\
        Final hidden state & whiten $+\,\ell_2$ (used) & 1{,}908 & 0.3 & 0.00 & 0.94 & 0.00 & 98.4 & 31.9 \\
        \bottomrule
    \end{tabular*}
    \endgroup
\end{table}

\subsection{Language interface configuration}
\label{app:interface}

\paragraph{Backbone and global memory.}
The language backbone is Qwen3-1.7B, with 28 decoder blocks and hidden size
$d_L=2048$ \citep{qwen3report}. Both routes use the same contextual EHR
sequence $\mathbf{H}$. The global resampler has four layers and 16 attention
heads, initialized from $K_g=64$ learned queries. At each layer, these queries
cross-attend to all valid encoder states and pass through a feed-forward block.
The resulting $64\times2048$ memory replaces the designated \texttt{<EMR>}
placeholder with continuous embeddings and also supplies the keys and values
of global cross-attention. The inserted embeddings may attend bidirectionally
to one another, while ordinary text attention remains causal.

\paragraph{Local projection and selector.}
The local projection and salience scorer are
\begin{align}
    \mathbf{c}_t &= \operatorname{LN}_{o}\!\left(
      \mathbf{W}_{\ell}\operatorname{LN}_{i}(\mathbf{H}_t)+\mathbf{b}_{\ell}
      \right),\\
    s_t &= \mathbf{w}_2^\top
      \operatorname{GELU}\!\left(\mathbf{W}_1
      \operatorname{LN}_{s}(\mathbf{c}_t)+\mathbf{b}_1\right)+b_2.
    \label{eq:app_local_projection}
\end{align}
The selector MLP has hidden width 1,024. Hard selection uses $K_{\ell}=256$
during concept alignment and $K_{\ell}=64$ in every downstream stage; when
fewer eligible events are available, all candidates are retained. Local
vectors are available through local cross-attention only and are not inserted
as ordinary LLM tokens. The selector does not depend on the downstream
instruction $q$, so it produces one ranking shared across tasks. It is learned
during concept alignment, adapted during the soft phase of rationale SFT
(Appendix~\ref{app:downstream}), and fixed thereafter.

\paragraph{Cross-attention placement and gates.}
Each route-specific cross-attention module is applied after its corresponding
original decoder block. Using zero-indexed blocks, the global modules are
attached at $\mathcal{S}_{\mathrm{glob}}=\{12,14,16,18,20,22\}$ and the local
modules at $\mathcal{S}_{\mathrm{loc}}=\{23,25,27\}$. Each module has one
learned scalar gate $\gamma_l^r$, initialized to $-2$, so its initial residual
weight is $\sigma(\gamma_l^r)\approx0.119$. The global interface is therefore
available before the final local injections. These depths are an empirical
configuration choice.

\subsection{Alignment objectives and training schedule}
\label{app:alignment}

Table~\ref{tab:app_training_stages} specifies trainable components throughout
the pipeline. The pretrained EHR encoder and base LLM are frozen in both
alignment stages. Direct prediction and rationale SFT are separate downstream
branches initialized from the aligned interface.

\begin{table}[t]
    \caption{Optimization stages. Global and local cross-attention include their
    scalar gates.}
    \label{tab:app_training_stages}
    \centering
    \small
    \begin{tabular}{p{0.18\linewidth}p{0.29\linewidth}p{0.43\linewidth}}
        \toprule
        Stage & Supervision & Trainable components \\
        \midrule
        Semantic encoder & masked concepts & EHR encoder and input projections \\
        Global alignment & BHC generation & resampler, global cross-attention \\
        Local alignment & concept descriptions + BHC replay & projection,
        selector, local cross-attention; local cross-attention only after hard
        selection \\
        Direct prediction & binary outcome & global and local cross-attention \\
        Rationale SFT, soft & teacher rationale + answer & base LLM, resampler,
        global and local cross-attention, local projection, selector \\
        Rationale SFT, hard & teacher rationale + answer & base LLM, resampler,
        global and local cross-attention \\
        GRPO & rule-based reward + answer loss & base LLM, global and local
        cross-attention \\
        \bottomrule
    \end{tabular}
\end{table}

\paragraph{Global trajectory alignment.}
The model receives a structured hospital trajectory and generates its
corresponding brief hospital course (BHC). We disable the local interface and
use the response-only next-token objective
\begin{equation}
    \mathcal{L}_{\mathrm{glob}}
      =-\sum_{j}\log p_{\theta}(b_j\mid b_{<j},q_{\mathrm{BHC}},
      \mathbf{M}^{\mathrm{glob}}),
    \label{eq:app_global_alignment}
\end{equation}
updating only the resampler and global cross-attention modules. Discharge text
provides an alignment target, not an input to downstream prediction.

\paragraph{Local concept alignment.}
We replace $5\%$ of eligible concepts by \texttt{[MASK]}, up to 32 positions
per example, and ask the model to generate descriptions for up to eight masked
concepts. Concept-description examples and BHC replay are mixed at rates $0.4$
and $0.6$, respectively. The EHR encoder, global route, and base LLM remain
frozen; the local projection, selector, and local cross-attention are initially
trainable. After the forced-selection phase, masked positions are excluded
from the selector's candidate set so the local route cannot simply forward
the representation at a masked slot.

The three phases are indexed by optimizer updates $u$:
\begin{equation}
\underbrace{\text{forced masked positions}}_{0\leq u<1000}
\quad\longrightarrow\quad
\underbrace{\text{soft selection},\;\tau_s:1.0\rightarrow0.3}_{1000\leq u<5000}
\quad\longrightarrow\quad
\underbrace{\text{hard top-}256}_{u\geq5000}.
\label{eq:selector_curriculum}
\end{equation}
In the first phase, masked positions are placed in local memory and binary
cross-entropy trains the selector to recognize them. In the soft phase, the
candidate memory remains full length but each vector is multiplied by
$\alpha_t=\operatorname{softmax}(s_t/\tau_s)$, allowing the generation loss to
reach the selector. An auxiliary squared budget loss, with weight $0.01$,
encourages the mean selection probability
$\frac{1}{|\mathcal{V}|}\sum_{t\in\mathcal{V}}\sigma(s_t)$ over valid candidates
$\mathcal{V}$ to approach $0.1$. At the transition to hard selection, the
projection and selector are frozen because the top-$k$ indices are
non-differentiable; local cross-attention continues to adapt on the same hard
route used downstream.

\subsection{Route specialization and the local memory budget}
\label{app:route_probe}
\providecommand{\tbd}{\textsf{TBD}}

\paragraph{Setup.}
We probe the final Stage-2 checkpoint, the one used by every downstream stage. Its early-stopping
copy was selected on BHC loss alone, early in training, before most of the local alignment. Each
probe evaluates the held-out alignment split (27{,}076 admissions of 12{,}835 validation patients)
under hard selection, with both routes open and with the local route, the global route, or both
switched off by forcing the corresponding gate logits to $-30$ ($\sigma\approx0$). Two validation targets are used: BHC
generation, which is the global alignment task, and concept-description generation with fixed
masked positions, which are excluded from the local candidates as during training. To choose the
downstream budget, the local memory is truncated to $K_\ell\in\{256,128,64\}$ at inference without
retraining. Switching the local route off, alone or together with the global route, removes the
local memory entirely, so those passes do not depend on $K_\ell$
(Table~\ref{tab:stage2_route_probe}).

\paragraph{The two routes specialize.}
Removing the local route raises concept-description loss by more than an order of magnitude
(0.280 to 3.311, $+1084\%$) but leaves BHC loss unchanged (1.794 to 1.792, $-0.1\%$). Removing the
global route roughly doubles both losses (BHC $+78.6\%$, concept description $+92.3\%$), and
removing both routes raises concept-description loss to 3.629. The local route is therefore
specialized to concept-level recovery, whereas the global route supports both objectives. The
pattern is not forced by the objectives: BHC replay during Stage~2 leaves the local route free to
contribute to BHC generation, yet BHC loss is unchanged without it. The trajectory summary also
does not substitute for access to individual events: with the local route off, the global route
alone cannot recover the masked concepts, which motivates keeping the local route for evidence
citation.

\paragraph{Choice of $K_\ell$.}
Truncating the local memory from 256 to 128 and 64 events raises concept-description loss by
1.7\% and 7.6\% with both routes open (0.280, 0.284, and 0.301) and by 2.9\% and 10.0\% with the
global route off, and leaves BHC loss unchanged (1.7937 vs.~1.7942). Even at $K_\ell=64$,
concept-description loss remains an order of magnitude below its value without the local route,
so the information the local route supplies is concentrated in the highest-ranked events. The budget also determines whether selection happens at all.
Table~\ref{tab:local_budget} shows that at $K_\ell=256$ the local memory would contain every
eligible event in 78.9\% of 24-h records and 58.5\% of readmission records, so the selector would
act as an identity on most inputs; the median 24-h record has only 96.5 eligible events. At
$K_\ell=64$, selection is active in 70.4\% of 24-h records and 79.7\% of readmission records. We
therefore align with $K_\ell=256$, which gives the selector and local cross-attention a broad
candidate set during training, and use $K_\ell=64$ in every downstream stage, where selection
remains a real choice at little cost in concept recovery.

\providecommand{\tbd}{\textsf{TBD}}
\begin{table}[t]
    \centering
    \caption{Route ablation after concept alignment (final Stage-2 checkpoint, hard selection): validation loss with
    the local and/or global cross-attention gates forced shut, and its change relative to the full model at the same
    $K_\ell$. Passes with the local route off do not read the local memory, so their values are shared across $K_\ell$
    ($\uparrow$: same as the $K_\ell{=}256$ row); \textemdash: not evaluated.}
    \label{tab:stage2_route_probe}
    \begingroup
    \footnotesize
    \setlength{\tabcolsep}{4pt}
    \begin{tabular*}{\linewidth}{@{\extracolsep{\fill}}llrrrr@{}}
        \toprule
        Validation target & $K_\ell$ & Full & Local off & Global off & Both off \\
        \midrule
        Concept description & 256 & 0.2797 & 3.3105 ($+$1083.5\%) & 0.5378 ($+$92.3\%) & 3.6285 \\
                            & 128 & 0.2844 & $\uparrow$ & 0.5536 ($+$94.7\%) & $\uparrow$ \\
                            & 64  & 0.3010 & $\uparrow$ & 0.5913 ($+$96.5\%) & $\uparrow$ \\
        \addlinespace[2pt]
        BHC generation      & 256 & 1.7937 & 1.7922 ($-$0.1\%) & 3.2044 ($+$78.6\%) & \textemdash \\
                            & 64  & 1.7942 & $\uparrow$ & \textemdash & \textemdash \\
        \bottomrule
    \end{tabular*}
    \endgroup
\end{table}

\begin{table}[t]
    \centering
    \caption{Local budget on the downstream test inputs (6{,}804 24-h and 7{,}566 readmission records; mortality and
    long length of stay share the 24-h inputs): eligible events per record (non-padding positions of the encoder input,
    truncated at 2{,}048, a limit that 10.2\% of readmission records reach) and the share of records in which top-$K_\ell$
    selection is inactive because the record has at most $K_\ell$ eligible events. On average, $K_\ell=256/128/64$ selects
    120.1/88.9/56.2 events for the 24-h inputs and 169.1/103.9/58.8 for readmission.}
    \label{tab:local_budget}
    \begingroup
    \footnotesize
    \setlength{\tabcolsep}{4pt}
    \begin{tabular*}{\linewidth}{@{\extracolsep{\fill}}lrrrrr@{}}
        \toprule
        & \multicolumn{2}{c}{Eligible events} & \multicolumn{3}{c}{Selection inactive ($\leq K_\ell$ events)} \\
        \cmidrule(lr){2-3}\cmidrule(l){4-6}
        Task & Median & p90 & $K_\ell=256$ & $K_\ell=128$ & $K_\ell=64$ \\
        \midrule
        Mortality, long LOS (24\,h) & 96.5 & 603 & 78.9\% & 66.5\% & 29.6\% \\
        Readmission (full) & 194 & 2{,}048 & 58.5\% & 37.8\% & 20.3\% \\
        \bottomrule
    \end{tabular*}
    \endgroup
\end{table}

\subsection{Downstream supervision and answer scoring}
\label{app:downstream}

\paragraph{Direct prediction.}
The direct branch uses a task instruction and scores the answer slot without
generating a rationale. For candidate $a\in\{\texttt{Yes},\texttt{No}\}$, let
$(v_1^a,\ldots,v_{n_a}^a)$ be its answer tokens. We compute
\begin{equation}
    s_a=\frac{1}{n_a}\sum_{i=1}^{n_a}
      \log p_{\theta}(v_i^a\mid\mathbf{x},q,v_{<i}^a),\qquad
    p_{\theta}(y=1\mid\mathbf{x},q)=
    \frac{\exp s_{\mathrm{Yes}}}
         {\exp s_{\mathrm{Yes}}+\exp s_{\mathrm{No}}}.
    \label{eq:yes_probability}
\end{equation}
In the multi-task setting, one interface is shared across the three task instructions. Only global and
local cross-attention modules are adapted by next-token cross-entropy on the
answer continuation and its terminating token. The EHR encoder, resampler,
local projection and selector, and base LLM remain frozen. At evaluation,
Equation~\ref{eq:yes_probability} averages answer-token log-likelihoods only,
excluding the terminating token. Hard top-64 selection is used during both
training and evaluation.

\paragraph{Supervised rationale initialization.}
The reasoning branch starts from the aligned model rather than the direct
checkpoint. Approximately 2.5K examples, balanced across task and label, use
rationales generated by Qwen3.5-397B-A17B \citep{qwen35release}. The teacher
receives the gold outcome and a textual rendering of the observed EHR; its
outputs are label-conditioned, post-hoc rationales, not independent annotations
of clinical causality. After format and content checks, the student is trained
to reproduce the rationale and final answer while seeing only the structured
EHR interface and task instruction. The EHR encoder remains frozen throughout,
while the base LLM, resampler, and both cross-attention interfaces are updated.

SFT runs in two consecutive phases on the same data. In the \emph{soft phase},
the local memory contains all eligible events weighted by
$\alpha_t=\operatorname{softmax}(s_t/\tau_s)$ with $\tau_s=0.3$, as in the soft
phase of concept alignment, and the local projection and selector are also
trainable, so the rationale loss reaches the event ranking. No selector budget
loss is used, keeping the objective a pure rationale cross-entropy. The
\emph{hard phase} starts from the best soft-phase checkpoint and restores hard
top-64 selection; because top-$k$ indices pass no gradient, the projection and
selector are frozen again, and the remaining components continue to adapt to
the memory used by GRPO and evaluation. Both phases use learning rate
$10^{-5}$ for the interface and $10^{-6}$ for the base LLM, weight decay $0.01$,
a cosine schedule with 5\% warmup restarted at the start of each phase, and at
most ten epochs with early stopping (patience 2, minimum improvement 0.005) on
the chain cross-entropy of a held-out 10\% split. The GRPO policy and its
reference are initialized from the best hard-phase checkpoint.

\subsection{Reward implementation and group relative policy optimization}
\label{app:reward}

\paragraph{Sampling and reward terms.}
Starting from the SFT checkpoint, the current run samples $G=4$ completions per
patient--task prompt, with temperature $0.8$, nucleus threshold $0.95$, and
token-sampling top-$k=20$. Generation is capped at 1,536 new tokens and stops
at the end of the answer. For a completion $o_j$, the reward is
\begin{equation}
    R(o_j)=r_{\mathrm{ans}}
      +0.10r_{\mathrm{fmt}}
      +0.15r_{\mathrm{evid}}
      -r_{\mathrm{len}},
    \label{eq:app_reward}
\end{equation}
where $r_{\mathrm{ans}}=1$ for a correct answer and $-1$ otherwise. The format
reward is $r_{\mathrm{fmt}}=2q-1$, where $q$ is the fraction of four structural
checks the response passes: a single reasoning block, an evidence heading
followed by at least one bullet, a reasoning heading followed by at least one
sentence, and exactly one final-answer tag closing the response. $q$ is reduced
by $0.2$, with a floor of zero, when the response contains an explicit event
identifier (\texttt{E} followed by three or more digits), which is not part of
the student output format. The length penalty is
\begin{equation}
    r_{\mathrm{len}}=\min\!\left(0.1,\,0.0005\max(0,|o_j|-768)\right).
    \label{eq:app_length_penalty}
\end{equation}
A missing or invalid final answer, or a detected reference to future outcome
information, overrides Equation~\ref{eq:app_reward} with $-1.5$, below the
lowest valid total of $-1.35$. The leakage check consists of patterns that
require outcome language in the past tense or an explicit reference to the
label: statements that the patient died, survived, was discharged, or was
readmitted; statements that no readmission occurred within 30 days; references
to the actual or observed length of stay; and references to a gold or provided
answer or label. Prospective statements such as ``high risk of death'' are not
flagged.

\paragraph{Evidence parsing and candidate events.}
Responses are parsed strictly, without repair. Evidence items are the bullet
lines (beginning with ``-'' or ``*'') between the heading ``Relevant clinical
evidence'' and the next ``Reasoning'', ``Analysis'', or ``Uncertainty'' heading.
An optional role tag (\texttt{risk\_factor}, \texttt{protective\_factor},
\texttt{counterevidence}, \texttt{uncertainty}, \texttt{supports\_yes}, or
\texttt{supports\_no}) is stripped and not scored. The candidate events are all
non-special, non-padding positions of the same truncated input window the model
receives ($T_{\max}=2048$; Appendix~\ref{app:encoder}), each represented by its
typed short description. Candidates therefore include events outside local
memory.

\paragraph{Lexical matcher.}
Text is lower-cased and split into alphanumeric tokens that begin with a letter.
Type prefixes (``condition:'', ``lab:'', and so on), single characters, function
words, and six generic words (\emph{patient}, \emph{clinical}, \emph{evidence},
\emph{history}, \emph{risk}, \emph{factor}) are removed, which gives a
content-token set $T(\cdot)$. Token statistics come from the $N=20{,}561$ typed
descriptions of the model vocabulary. With $\operatorname{df}(t)$ the number of
descriptions containing token $t$,
$\operatorname{idf}(t)=\log\frac{N+1}{\operatorname{df}(t)+1}+1$, and $t$ is an
\emph{anchor} if $\operatorname{df}(t)\le0.02N$ (411 descriptions). The
direction words \emph{very}, \emph{high}, \emph{low}, and \emph{normal} are
never anchors. Anchors are therefore derived from the vocabulary rather than
from an authored list of medical terms; the most frequent content tokens are
drug-exposure and dosage words such as \emph{exposure}, \emph{mg}, and
\emph{oral}. For a bullet $p$ and a candidate description $d$, the matcher
applies four rules.
\begin{enumerate}
    \item \emph{Direction.} If $p$ states \emph{very high}, \emph{high},
    \emph{very low}, \emph{low}, or \emph{normal}, $d$ must state the same
    direction. A bullet without a direction may cite a directional event.
    \item \emph{Qualifiers.} $p$ and $d$ must assert the same negation
    qualifiers (``no evidence of'', ``no history of'', ``no sign(s) of'',
    ``negative for'', ``deny/denies/denied'', ``ruled out'', ``absence of'',
    ``not present'') and the same family-history attribution. ``Without'' and
    ``not'' are excluded because they act as specifiers in OMOP concept names
    (``\ldots\ without hepatic coma'', ``not otherwise specified''). Across
    59,929 logged evidence phrases, ``without'' caused 30 rejections, all of
    legitimate citations, whereas the retained qualifiers caused none.
    \item \emph{Score.} The IDF-weighted Dice overlap is
    \begin{equation}
        s(p,d)=\frac{2\sum_{t\in T(p)\cap T(d)}\operatorname{idf}(t)}
        {\sum_{t\in T(p)}\operatorname{idf}(t)+\sum_{t\in T(d)}\operatorname{idf}(t)}.
        \label{eq:app_match_score}
    \end{equation}
    \item \emph{Support.} $p$ is supported by $d$ if $T(p)\subseteq T(d)$ or
    $T(d)\subseteq T(p)$, or if every anchor in $T(p)$ occurs in $T(d)$ and
    $s(p,d)\ge0.55$.
\end{enumerate}
Each bullet is assigned the candidate that is largest in the order (supported,
$s$, number of shared anchors); a bullet that no candidate supports is
unsupported. The anchor rule is precision-first: ``acute on chronic renal
failure'' cannot cite ``acute on chronic heart failure'' even though most
tokens overlap, because the anchor \emph{renal} is unmatched. The matcher is
lexical rather than ontological, which leaves two known failure modes. A
paraphrase that introduces a rare word absent from the description, such as
``kidney failure'' for a renal-failure event, is not matched, so support is
conservative for paraphrased findings. Conversely, containment in either
direction accepts under-specified bullets: a bare laboratory name such as
``Sodium'' is supported by the event ``Sodium [Moles/volume] in Serum or
Plasma, low'' and receives that event's credit.

\paragraph{Distinct-concept credit.}
$f(c)$ is the fraction of GRPO training records, pooled over the three tasks
and computed on the untruncated inputs, that contain concept $c$. The $i$-th
bullet receives $\xi_i=1-f(c)$ if it is the first supported citation of $c$ in
the response, and $\xi_i=0$ if it is unsupported, cites a concept already
credited in the response, or matches a metadata description (age, sex, gender,
visit, or discharge). Metadata is excluded because each individual value, for
example one of 126 age buckets, is rare by construction and would otherwise
earn nearly full credit. With $m$ bullets and $m_0=4$,
\begin{equation}
    S=\frac{1}{\max(m,m_0)}\sum_{i=1}^{m}\xi_i,
    \qquad
    r_{\mathrm{evid}}=2S-1\quad(m>0),
    \qquad
    r_{\mathrm{evid}}=-1\quad(m=0).
    \label{eq:app_evidence_reward}
\end{equation}
Two development runs motivated this form. In a run that credited every
supported bullet, rollouts with ten or more duplicated bullets averaged a total
reward of $0.63$, against $0.15$ for rollouts with fewer than three, because
repeating one supported finding diluted the unsupported bullets. First-citation
credit removes this gain. In a run whose denominator was $m$ alone, the mean
number of bullets fell from $10.96$ to $5.63$ over 1,880 steps while weighted
precision rose from $0.18$ to $0.38$. After about 1,100 steps the number of
distinct supported findings also fell, from $3.09$ to $2.64$. The floor $m_0$ removes the gain from shortening a list below $m_0$ bullets.
Above $m_0$, dropping a bullet whose credit is below the current average still
raises $S$, so we report distinct supported findings together with precision;
in the final run, GRPO shortened responses from 10.1 to 7.8 bullets while
distinct supported findings rose (Section~\ref{sec:evidence}). Credit is the same whether the matched event lies inside
or outside local memory; that distinction is logged only for analysis.

\paragraph{Evidence-sufficiency loss.}
For at most two responses per group, rotated across steps so that every
response position is reached, the response is reduced to its evidence heading
and bullets, the reasoning prose and answer are removed, and the answer prefix
is appended. The EHR is then silenced: the resampled global memory tokens are
zeroed and both cross-attention routes are bypassed. The cross-entropy of the
gold answer token under this input is added to the loss with weight $0.2$ and
no class weighting, so its gradient reaches the language model only through the
text path. Responses without bullets are skipped because the evidence term
already assigns them $-1$. Only the bullets are shown because, with the full
rationale visible, the loss can be minimized by stating the conclusion in the
prose. As a reward, $2p_{\mathrm{gold}}-1$ would be mean-subtracted within each
group. On logged rollouts its within-group slope in the number of bullets was
$-0.024$ for negative-label prompts and $+0.029$ for positive-label prompts, so
with about 72\% negative prompts the reward form would pay for citing less. As a
loss it is a proper scoring rule with no group baseline to exploit. The
bullets-only answer probability is also logged as a training diagnostic but is
not part of the reward.

\paragraph{Policy update.}
Rewards are standardized within each group and advantages are clipped:
\begin{equation}
    A_j=\operatorname{clip}\!\left(
      \frac{R(o_j)-\overline{R}}
      {\operatorname{sd}(R(o_1),\ldots,R(o_G))+\epsilon},-5,5\right),
    \label{eq:grpo_advantage}
\end{equation}
with zero advantage when the group variance is zero. The GRPO loss combines the
clipped policy-ratio surrogate \citep{shao2024deepseekmath}, a reference-policy
KL penalty with coefficient $0.05$, the answer loss, and the
evidence-sufficiency loss above. The answer loss is the cross-entropy of the
gold answer token at the answer position of each sampled response with a valid
answer, with coefficient $0.2$ and no class weighting. The KL term is omitted
for zero-variance groups, whose only remaining gradient would otherwise pull the
policy toward the reference model. In the current configuration there is one
inner update, and old-policy log-probabilities are detached current
log-probabilities. The importance ratio is therefore one on that update and
clipping is inactive. Operationally, this run uses a group-normalized
policy-gradient update with KL and auxiliary-loss regularization. The EHR encoder,
resampler, local projection, and selector remain frozen; the base LLM and both
cross-attention interfaces are updated.

\subsection{Evidence intervention and route probes}
\label{app:intervention}

Matcher support alone does not establish that a cited event affected model
behavior \citep{jacovi2020faithfulness}. We define a pre-generation intervention
to assess this dependence. For each example, we first decode the full
rationale and answer greedily, map citations corresponding to selected
local-memory concepts, and record $p_{\mathrm{Yes}}$. We then mask every mapped
local slot before generating the first output token and regenerate the complete
output. The control intervention masks the same number of uncited selected
slots, matched as closely as possible by $1-f(c)$, because removing inputs
without retraining can itself change model behavior \citep{hooker2019roar}. We compare three quantities
between the cited and control interventions: persistence, the share of the
masked concepts that the regenerated rationale still cites; the change in
$p_{\mathrm{Yes}}$ and in the predicted label relative to the full chain; and
AUROC, with the answer read either under the same mask or with the local route
switched off, which removes direct access to local memory; masked evidence can
then reach the answer only through the regenerated text or the global memory.

\paragraph{Sample.} The intervention is applied to the first 500 test cases of
each task and to 1{,}000 further readmission cases, added to test whether a
readmission effect seen in the first 500 replicates. Under greedy decoding,
2{,}280 of the 2{,}500 chains reach the answer anchor, and 2{,}062 of them cite
at least one concept held in local memory (mortality 453 with 17 positives, long
LOS 441 with 152, readmission 1{,}168 with 75). A chain cites 3.2 such concepts
on average, occupying 4.3 of the slots of about 44 distinct concepts in local
memory; cited and control concepts have similar informativeness (0.826 vs.~0.833, against 0.715 for all local concepts).

\paragraph{Results.} Masking the cited slots lowers persistence by 0.017
[0.007, 0.028] relative to the control mask, most clearly on readmission
(Table~\ref{tab:mediation}), so the local route contributes to which findings are
cited. The answer is no more sensitive to the cited mask than to the control
mask. Before extending the readmission sample, we fixed the primary test as the
cited-minus-control AUROC on readmission with the local route off, evaluated on
the 1{,}000 added cases, with all 1{,}500 cases as a secondary test. A difference
of $-0.010$ [$-0.027$, $0.004$] in the first 500 cases did not replicate
($+0.002$ [$-0.012$, $0.019$]; pooled $-0.003$ [$-0.013$, $0.007$]).

\begin{table}[!htbp]
    \centering
    \caption{Cited-slot masking on MIMIC-IV test cases whose greedy chain cites concepts held in local memory.
    Persistence: share of the masked concepts (cited in the full chain) that the regenerated rationale still cites.
    $|\Delta p_{\mathrm{Yes}}|$ and flips: change of the answer probability and of the predicted label relative to the
    full chain. AUROC: cited minus control, with the answer read under the mask or with the local route switched off.
    Brackets: two-sided 95\% paired bootstrap intervals. AUROC refers to this greedy subset and is not comparable with
    Table~\ref{tab:outcome_prediction}.}
    \label{tab:mediation}
    \begingroup
    \footnotesize
    \setlength{\tabcolsep}{3pt}
    \newcommand{\wci}[3]{\begin{tabular}[t]{@{}c@{}}$#1$\\[-1pt]{\scriptsize[$#2$, $#3$]}\end{tabular}}
    \begin{tabular*}{\linewidth}{@{\extracolsep{\fill}}lcccc@{}}
        \toprule
        & Mortality & Long LOS & Readmission & All \\
        \midrule
        Cases (positives) & 453 (17) & 441 (152) & 1{,}168 (75) & 2{,}062 (244) \\
        Persistence, cited mask & 0.874 & 0.885 & 0.851 & 0.863 \\
        Persistence, control mask & 0.885 & 0.889 & 0.875 & 0.880 \\
        Persistence, cited $-$ control & \wci{-0.011}{-0.033}{0.010} & \wci{-0.005}{-0.026}{0.018} & \wci{-0.024}{-0.040}{-0.009} & \wci{-0.017}{-0.028}{-0.007} \\
        $|\Delta p_{\mathrm{Yes}}|$, cited / control & 0.007 / 0.007 & 0.021 / 0.021 & 0.012 / 0.012 & 0.013 / 0.013 \\
        Label flips (\%), cited / control & 1.1 / 1.3 & 1.8 / 2.9 & 0.4 / 0.3 & 0.9 / 1.1 \\
        AUROC, cited $-$ control, masked & \wci{-0.001}{-0.008}{0.004} & \wci{0.001}{-0.005}{0.008} & \wci{-0.009}{-0.023}{0.005} & \wci{-0.003}{-0.009}{0.002} \\
        AUROC, cited $-$ control, local off & \wci{-0.005}{-0.013}{0.001} & \wci{0.002}{-0.005}{0.009} & \wci{-0.003}{-0.014}{0.007} & \wci{-0.002}{-0.007}{0.002} \\
        \bottomrule
    \end{tabular*}
    \endgroup
\end{table}

The intervention measures output sensitivity to cited local representations,
not causal effects of the corresponding clinical events. Global memory and
unmasked local slots remain available, so a small intervention effect can
reflect redundant evidence. A cross-attention gate probe also differs from
complete route removal: in a global-cross-attention-off probe, the 64 inserted
global embeddings remain in the LLM sequence. Such a probe measures the
incremental effect of global cross-attention and cannot be described as removing
all global information.

\section{Data and cohort construction}
\label{app:data}

\begin{figure}[p]
    \centering
    \includegraphics[width=0.8\linewidth,trim=21.5 30.3 603.2 36.5,clip]{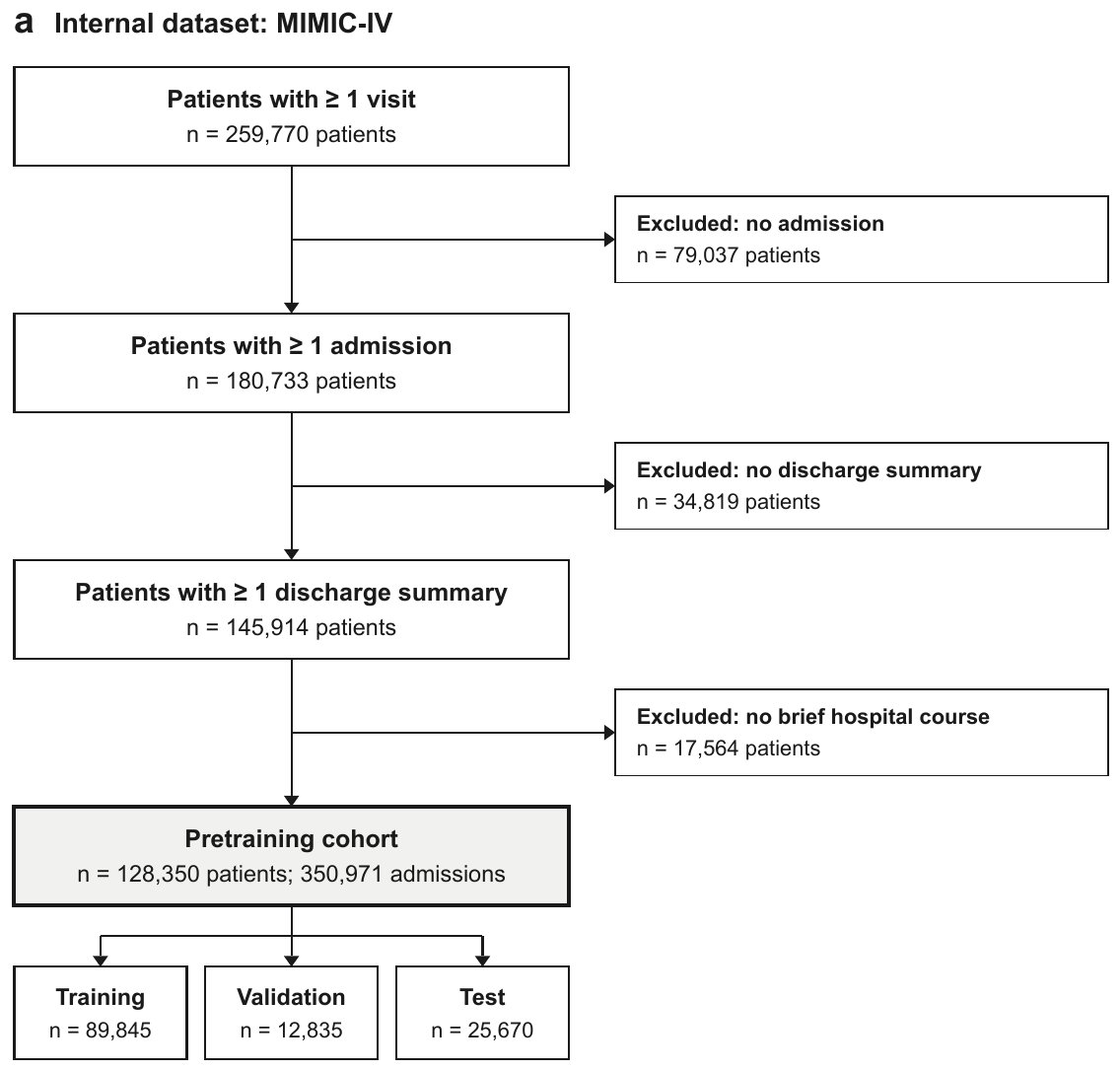}\\[6pt]
    \includegraphics[width=0.8\linewidth,trim=598.7 104.9 26.0 33.3,clip]{figures/cohort_fig.pdf}
    \caption{Cohort construction. (a) Internal MIMIC-IV cohort (OMOP CDM format),
    split by patient. (b) EHRSHOT external cohort; inpatient visits that start
    within one day of the previous discharge are merged into one admission.
    Task-level sample sizes are given in Table~\ref{tab:task_cohorts}.}
    \label{fig:cohort}
\end{figure}

\paragraph{Internal cohort.}
Our main dataset is MIMIC-IV v2.2 \citep{johnson2023mimiciv}, converted to the
OMOP CDM format; it contains 259,770 patients with at least one visit and
431,231 hospital admissions. Discharge summaries come from MIMIC-IV-Note v2.2
\citep{johnson2023mimicivnote}, and BHC targets come from MIMIC-IV-Ext-BHC
\citep{aali2025bhc}, which provides the brief hospital course section of
270,033 discharge notes. Figure~\ref{fig:cohort}a shows the cohort. Of the
patients with at least one visit, 180,733 have a hospital admission, 145,914
have a discharge summary, and 128,350 have at least one admission with an
extractable BHC. These 128,350 patients form the cohort. They are split by
patient into training (89,845), validation (12,835), and test (25,670) sets,
a 70/10/20 split with a fixed seed.

\paragraph{Use of the split.}
Every stage fits its parameters on training patients only: EHR-BERT
pretraining, both alignment stages (189,115 training and 27,076 validation
admission--BHC pairs), rationale SFT, GRPO, and all supervised baselines. The
teacher rationales for SFT are generated from training records only.
Checkpoints are chosen per stage: EHR-BERT pretraining and global alignment by
validation loss; the supervised baselines and the direct GLoC-EHR variants by
validation AUROC (Section~\ref{sec:outcome_comparison}); rationale SFT by early
stopping on a held-out 10\% of its training examples
(Appendix~\ref{app:downstream}); and concept alignment and GRPO use their final
checkpoints (Appendices~\ref{app:route_probe} and~\ref{app:reasoning_snapshot}).
The downstream tasks inherit the split, so test records are used neither to
fit model parameters nor to select checkpoints.
bin thresholds were derived from the full cohort before splitting. Neither uses
outcome labels, and refitting the bins on training patients alone would change
the value bin of 0.35\% of measurement events.

\paragraph{Downstream tasks.}
Index admissions are inpatient visits, with OMOP visit concept Inpatient Visit
or Emergency Room and Inpatient Visit, and every eligible admission of a
patient is used. Mortality and long length of stay share one set of
admissions: stays of at least 48 hours, excluding patients who died before the
input cutoff. The input contains the events recorded in the first 24 hours of
the admission, with discharge information removed. Mortality is death during
the admission, and long length of stay is a stay longer than 7 days.
Thirty-day readmission uses the full index admission with its
discharge-disposition tokens removed and excludes admissions that end in
death. It is positive when the next inpatient admission begins within 30 days
of discharge; a negative additionally requires that the data extend at least
30 days beyond discharge. Table~\ref{tab:task_cohorts} gives the resulting
sizes and prevalences.

\paragraph{Serialized record length.}
The token counts in Section~\ref{sec:introduction} serialize the same event
sequences that the models receive, without deduplication, using the typed short
concept descriptions, and add the direct-answer instruction (about 134 tokens).
Tokens are counted with the Qwen3-1.7B tokenizer over all training, validation,
and test cases of each task.

\begin{table}[!hb]
    \centering
    \caption{Downstream sample sizes (admissions) and outcome prevalence. Mortality
    and long length of stay (LOS) share the same admissions and 24-h inputs.
    EHRSHOT is used for evaluation only.}
    \label{tab:task_cohorts}
    \begingroup
    \footnotesize
    \setlength{\tabcolsep}{2.5pt}
    \begin{tabular*}{\linewidth}{@{\extracolsep{\fill}}llrrrr@{}}
        \toprule
        & & \multicolumn{3}{c}{MIMIC-IV} & \\
        \cmidrule(lr){3-5}
        Task & Input & Training & Validation & Test & EHRSHOT \\
        \midrule
        In-hospital mortality & First 24\,h & 24,876 (3.19\%) & 3,549 (3.18\%) & 6,804 (3.15\%) & 9,922 (2.75\%) \\
        Long LOS ($>$7\,d) & First 24\,h & 24,876 (37.06\%) & 3,549 (37.70\%) & 6,804 (36.67\%) & 9,922 (31.14\%) \\
        30-day readmission & Full admission & 27,420 (7.55\%) & 3,890 (7.46\%) & 7,566 (6.89\%) & 12,342 (23.70\%) \\
        \bottomrule
    \end{tabular*}
    \endgroup
\end{table}

\paragraph{External cohort.}
The EHRSHOT release \citep{wornow2023ehrshot} we use contains 6,731 patients
in the OMOP CDM; 6,710 have at least one visit and 4,498 have at least one
inpatient visit (Figure~\ref{fig:cohort}b). EHRSHOT records one hospital stay
as several consecutive visits: 60.9\% of consecutive inpatient visits start
within one day of the previous discharge. For readmission, such visits are
merged into one admission (40,627 visits into 12,699 admissions), and the
longest fragment serves as the index; without merging, fragments of the same
stay would count as readmissions and raise the prevalence to 0.39. Mortality
and long length of stay use visit records directly, because the 48-hour
minimum already excludes short fragments. The task definitions are otherwise
identical to the internal ones, and all models are applied without
retraining.

\section{Additional outcome-prediction details and scores}
\label{app:prediction_scores}

\paragraph{Tasks and evaluation.}
All models use the same patient-level MIMIC-IV splits
(Appendix~\ref{app:data}). Mortality and long length of stay
are predicted from the first 24 hours of the index admission (test $n=6{,}804$;
prevalence 3.15\% and 36.67\%), and 30-day readmission from the full index
admission (test $n=7{,}566$; prevalence 6.89\%). Supervised configurations are
selected by task-specific validation AUROC, or by validation macro AUROC over
the three tasks for multi-task models, and the selected configuration is
retrained with seeds 42--46. We report test means and sample SDs ($\mathrm{ddof}=1$).
Macro scores average the three tasks within each seed before aggregating
across seeds; for task-specific models they summarize separately selected
runs rather than a common configuration. Logistic-regression fits are
deterministic across seeds and are shown without SD. Bold and underlined
values mark the best and second-best distinct four-decimal means within each
column; they are descriptive and do not indicate statistical significance.

\paragraph{Zero-shot LLM baselines.}
Each record is serialized as a newline-separated list of concept names,
ordered from most to least recent, de-duplicated, and truncated to 8{,}192
tokens \citep{hegselmann2026ehrencoders}. The prompt follows the ClinicalBench
instruction \citep{chen2024clinicalbench} (``Answer 1 if yes, 0 if no''). The
predicted probability is the next-token probability of the answer at the
answer slot, renormalized over the two options. Direct prompts use greedy
decoding. For CoT, Llama-3.3-70B-Instruct decodes greedily (at most 1{,}536
tokens), and Qwen3-1.7B uses thinking mode ($T{=}0.6$, top-$p$ 0.95,
top-$k$ 20, at most 3{,}072 tokens). Chains that do not reach an answer are
closed with the answer prefix (forced closure); Table~\ref{tab:reasoning_coverage}
reports how often this occurs. The ``our prompt'' variants replace the
ClinicalBench instruction with the Yes/No evidence-then-answer format of
GLoC-EHR (Reasoning).

\paragraph{Count-based and sequence baselines.}
Count features contain 20{,}567 code counts. LR sweeps
$C\in\{10^{-4},\ldots,10\}$ with and without positive-class weighting (12
configurations); LightGBM and XGBoost sweep learning rate $\{0.05,0.1\}$,
tree size, and positive-class weighting (8 configurations each) with up to
3{,}000 boosting rounds and early stopping. GRU \citep{cho2014gru} and RETAIN
\citep{choi2016retain} sweep learning rate, hidden size, positive-class
weighting, and batch size (24 configurations each) over the same 2{,}048-event
inputs.

\paragraph{Frozen-vector baselines.}
$T_{4096}$ is the 4{,}096-dimensional Qwen3-Embedding-8B embedding of the
serialized record, $E_{\mathrm{CLS}}$ and $E_{\mathrm{mean}}$ are
768-dimensional frozen EHR-BERT CLS and mean-pooled vectors, and $C$ denotes
code counts. Seven feature sets ($E_{\mathrm{CLS}}$, $E_{\mathrm{mean}}$,
$T_{4096}$, $T_{4096}{+}E_{\mathrm{CLS}}$, $C{+}E_{\mathrm{CLS}}$,
$C{+}T_{4096}$, $C{+}T_{4096}{+}E_{\mathrm{CLS}}$) are paired with LR and
LightGBM heads, and $C{+}E_{\mathrm{CLS}}$ additionally with XGBoost, giving 15
feature--head combinations with the same sweeps as above. These are local
adaptations of the LLM-as-EHR-encoder approach of
\citet{hegselmann2026ehrencoders}, not reproductions of its architecture or
published results.

\paragraph{Fine-tuned models.}
EHR-BERT (ST) sweeps learning rate $\{3,5,10,20\}\times10^{-6}$ and label
sampling (natural or power-tempered with $\alpha=0.5$); EHR-BERT (MT) uses the
same grid with task-balanced sampling and one prediction head per task.
GLoC-EHR (ST) and (MT) sweep learning rate $\{1,2,3,5,10\}\times10^{-4}$ with
the same sampling options (task-balanced for MT), use hard top-64 local selection, and score only the
answer tokens renormalized over Yes/No (Appendix~\ref{app:downstream}). All
fine-tuned models use early stopping with patience 5.

Tables~\ref{tab:outcome_prediction_full} and~\ref{tab:outcome_prediction_full_auprc}
report AUROC and AUPRC for every model; AUPRC is more informative than AUROC
when positives are rare \citep{saito2015prc}.

\begin{table}[!htbp]
    \centering
    \caption{Full internal AUROC comparison on MIMIC-IV (test): mean $\pm$ SD over five seeds, or over five sampled chains of one run for GLoC-EHR (Reasoning); rows without SD are single runs or deterministic (LR). Best in bold, second underlined. ST: one model per task; MT: one model for all three tasks. $C$: code counts; $T_{4096}$: text embedding; $E_{\mathrm{CLS}}$/$E_{\mathrm{mean}}$: frozen EHR-BERT vectors; a slash names the classifier head.}
    \label{tab:outcome_prediction_full}
    \begingroup
    \footnotesize
    \setlength{\tabcolsep}{2.5pt}
    \renewcommand{\arraystretch}{1.1}
    \newcommand{\predscore}[2]{\ensuremath{#1{\scriptstyle\,\pm\,#2}}}
    \begin{tabular*}{\linewidth}{@{\extracolsep{\fill}}lcccc@{}}
        \toprule
        Method & Mortality (24\,h) & Long LOS (24\,h) & Readmission (full) & Macro (3 tasks) \\
        \midrule
        \multicolumn{5}{@{}l}{\textit{Zero-shot LLMs on serialized records (single run)}} \\
        Qwen3-1.7B (direct) & \ensuremath{0.8140} & \ensuremath{0.6062} & \ensuremath{0.5183} & \ensuremath{0.6461} \\
        Qwen3-1.7B (CoT) & \ensuremath{0.5488} & \ensuremath{0.5363} & \ensuremath{0.5052} & \ensuremath{0.5301} \\
        Qwen3-1.7B (CoT, our prompt) & \ensuremath{0.5203} & \ensuremath{0.5136} & \ensuremath{0.4923} & \ensuremath{0.5087} \\
        Llama-3.3-70B (direct) & \ensuremath{0.8268} & \ensuremath{0.6391} & \ensuremath{0.5269} & \ensuremath{0.6643} \\
        Llama-3.3-70B (CoT) & \ensuremath{0.6780} & \ensuremath{0.5193} & \ensuremath{0.5034} & \ensuremath{0.5669} \\
        Llama-3.3-70B (CoT, our prompt) & \ensuremath{0.6464} & \ensuremath{0.5004} & \ensuremath{0.5006} & \ensuremath{0.5491} \\
        \addlinespace[2pt]
        \multicolumn{5}{@{}l}{\textit{Count-based models}} \\
        LR & \ensuremath{0.9448} & \ensuremath{0.8463} & \ensuremath{0.7615} & \ensuremath{0.8509} \\
        LightGBM & \predscore{0.9527}{0.0013} & \predscore{0.8624}{0.0011} & \predscore{0.7758}{0.0028} & \predscore{0.8637}{0.0013} \\
        XGBoost & \predscore{0.9525}{0.0009} & \predscore{0.8620}{0.0004} & \predscore{0.7733}{0.0035} & \predscore{0.8626}{0.0010} \\
        \addlinespace[2pt]
        \multicolumn{5}{@{}l}{\textit{Sequence models}} \\
        GRU & \predscore{0.9232}{0.0023} & \predscore{0.8336}{0.0040} & \predscore{0.7400}{0.0097} & \predscore{0.8323}{0.0037} \\
        RETAIN & \predscore{0.9186}{0.0054} & \predscore{0.8341}{0.0046} & \predscore{0.7415}{0.0045} & \predscore{0.8314}{0.0030} \\
        \addlinespace[2pt]
        \multicolumn{5}{@{}l}{\textit{Frozen-vector heads}} \\
        $E_{\mathrm{CLS}}$ / LR & \ensuremath{0.9375} & \ensuremath{0.8397} & \ensuremath{0.7595} & \ensuremath{0.8455} \\
        $E_{\mathrm{CLS}}$ / LGBM & \predscore{0.9292}{0.0023} & \predscore{0.8385}{0.0043} & \predscore{0.7481}{0.0098} & \predscore{0.8386}{0.0042} \\
        $E_{\mathrm{mean}}$ / LR & \ensuremath{0.9285} & \ensuremath{0.8375} & \ensuremath{0.7600} & \ensuremath{0.8420} \\
        $E_{\mathrm{mean}}$ / LGBM & \predscore{0.9125}{0.0050} & \predscore{0.8389}{0.0018} & \predscore{0.7556}{0.0034} & \predscore{0.8357}{0.0016} \\
        $T_{4096}$ / LR & \ensuremath{0.9148} & \ensuremath{0.8175} & \ensuremath{0.7677} & \ensuremath{0.8333} \\
        $T_{4096}$ / LGBM & \predscore{0.9062}{0.0018} & \predscore{0.8196}{0.0006} & \predscore{0.7521}{0.0032} & \predscore{0.8260}{0.0016} \\
        $T_{4096}{+}E_{\mathrm{CLS}}$ / LR & \ensuremath{0.9400} & \ensuremath{0.8440} & \ensuremath{0.7765} & \ensuremath{0.8535} \\
        $T_{4096}{+}E_{\mathrm{CLS}}$ / LGBM & \predscore{0.9385}{0.0008} & \predscore{0.8493}{0.0018} & \predscore{0.7620}{0.0043} & \predscore{0.8499}{0.0015} \\
        $C{+}E_{\mathrm{CLS}}$ / LR & \ensuremath{0.9481} & \ensuremath{0.8523} & \ensuremath{0.7653} & \ensuremath{0.8553} \\
        $C{+}E_{\mathrm{CLS}}$ / LGBM & \predscore{0.9497}{0.0016} & \predscore{0.8580}{0.0010} & \predscore{0.7577}{0.0060} & \predscore{0.8552}{0.0016} \\
        $C{+}E_{\mathrm{CLS}}$ / XGB & \predscore{0.9520}{0.0010} & \predscore{0.8576}{0.0006} & \predscore{0.7677}{0.0029} & \predscore{0.8591}{0.0008} \\
        $C{+}T_{4096}$ / LR & \ensuremath{0.9181} & \ensuremath{0.8225} & \ensuremath{0.7685} & \ensuremath{0.8364} \\
        $C{+}T_{4096}$ / LGBM & \predscore{0.9396}{0.0042} & \predscore{0.8504}{0.0015} & \predscore{0.7632}{0.0039} & \predscore{0.8510}{0.0014} \\
        $C{+}T_{4096}{+}E_{\mathrm{CLS}}$ / LR & \ensuremath{0.9375} & \ensuremath{0.8459} & \ensuremath{0.7764} & \ensuremath{0.8533} \\
        $C{+}T_{4096}{+}E_{\mathrm{CLS}}$ / LGBM & \predscore{0.9449}{0.0027} & \predscore{0.8589}{0.0013} & \predscore{0.7625}{0.0068} & \predscore{0.8554}{0.0017} \\
        \addlinespace[2pt]
        \multicolumn{5}{@{}l}{\textit{Fine-tuned, task-specific (ST)}} \\
        EHR-BERT (ST) & \predscore{\mathbf{0.9607}}{0.0018} & \predscore{\underline{0.8701}}{0.0008} & \predscore{0.7817}{0.0057} & \predscore{0.8708}{0.0020} \\
        GLoC-EHR (ST) & \predscore{0.9561}{0.0017} & \predscore{\mathbf{0.8740}}{0.0018} & \predscore{\mathbf{0.8075}}{0.0015} & \predscore{\mathbf{0.8792}}{0.0008} \\
        \addlinespace[2pt]
        \multicolumn{5}{@{}l}{\textit{Fine-tuned, multi-task (MT)}} \\
        EHR-BERT (MT) & \predscore{\underline{0.9575}}{0.0017} & \predscore{0.8672}{0.0019} & \predscore{0.7865}{0.0026} & \predscore{0.8704}{0.0013} \\
        GLoC-EHR (MT) & \predscore{0.9545}{0.0013} & \predscore{0.8671}{0.0030} & \predscore{0.7976}{0.0052} & \predscore{\underline{0.8731}}{0.0030} \\
        GLoC-EHR (Reasoning) & \predscore{0.9459}{0.0010} & \predscore{0.8469}{0.0004} & \predscore{\underline{0.8042}}{0.0018} & \predscore{0.8656}{0.0008} \\
        \bottomrule
    \end{tabular*}
    \endgroup
\end{table}

\begin{table}[!htbp]
    \centering
    \caption{Full internal AUPRC comparison on MIMIC-IV (test; prevalence 3.15\% / 36.67\% / 6.89\%): mean $\pm$ SD over five seeds, or over five sampled chains of one run for GLoC-EHR (Reasoning); rows without SD are single runs or deterministic (LR). Best in bold, second underlined. ST: one model per task; MT: one model for all three tasks. $C$: code counts; $T_{4096}$: text embedding; $E_{\mathrm{CLS}}$/$E_{\mathrm{mean}}$: frozen EHR-BERT vectors; a slash names the classifier head.}
    \label{tab:outcome_prediction_full_auprc}
    \begingroup
    \footnotesize
    \setlength{\tabcolsep}{2.5pt}
    \renewcommand{\arraystretch}{1.1}
    \newcommand{\predscore}[2]{\ensuremath{#1{\scriptstyle\,\pm\,#2}}}
    \begin{tabular*}{\linewidth}{@{\extracolsep{\fill}}lcccc@{}}
        \toprule
        Method & Mortality (24\,h) & Long LOS (24\,h) & Readmission (full) & Macro (3 tasks) \\
        \midrule
        \multicolumn{5}{@{}l}{\textit{Zero-shot LLMs on serialized records (single run)}} \\
        Qwen3-1.7B (direct) & \ensuremath{0.1203} & \ensuremath{0.4797} & \ensuremath{0.0718} & \ensuremath{0.2240} \\
        Qwen3-1.7B (CoT) & \ensuremath{0.0347} & \ensuremath{0.3849} & \ensuremath{0.0696} & \ensuremath{0.1631} \\
        Qwen3-1.7B (CoT, our prompt) & \ensuremath{0.0327} & \ensuremath{0.3732} & \ensuremath{0.0679} & \ensuremath{0.1579} \\
        Llama-3.3-70B (direct) & \ensuremath{0.0916} & \ensuremath{0.4552} & \ensuremath{0.0727} & \ensuremath{0.2065} \\
        Llama-3.3-70B (CoT) & \ensuremath{0.0480} & \ensuremath{0.3759} & \ensuremath{0.0693} & \ensuremath{0.1644} \\
        Llama-3.3-70B (CoT, our prompt) & \ensuremath{0.0439} & \ensuremath{0.3669} & \ensuremath{0.0689} & \ensuremath{0.1599} \\
        \addlinespace[2pt]
        \multicolumn{5}{@{}l}{\textit{Count-based models}} \\
        LR & \ensuremath{0.3969} & \ensuremath{0.7772} & \ensuremath{0.3221} & \ensuremath{0.4987} \\
        LightGBM & \predscore{0.4524}{0.0190} & \predscore{0.7966}{0.0021} & \predscore{0.3406}{0.0096} & \predscore{0.5299}{0.0102} \\
        XGBoost & \predscore{\underline{0.4612}}{0.0058} & \predscore{0.7980}{0.0010} & \predscore{0.3489}{0.0048} & \predscore{0.5361}{0.0021} \\
        \addlinespace[2pt]
        \multicolumn{5}{@{}l}{\textit{Sequence models}} \\
        GRU & \predscore{0.2727}{0.0099} & \predscore{0.7512}{0.0048} & \predscore{0.2945}{0.0110} & \predscore{0.4395}{0.0022} \\
        RETAIN & \predscore{0.3377}{0.0311} & \predscore{0.7579}{0.0064} & \predscore{0.2991}{0.0031} & \predscore{0.4649}{0.0109} \\
        \addlinespace[2pt]
        \multicolumn{5}{@{}l}{\textit{Frozen-vector heads}} \\
        $E_{\mathrm{CLS}}$ / LR & \ensuremath{0.3795} & \ensuremath{0.7585} & \ensuremath{0.3157} & \ensuremath{0.4845} \\
        $E_{\mathrm{CLS}}$ / LGBM & \predscore{0.3479}{0.0112} & \predscore{0.7526}{0.0070} & \predscore{0.3301}{0.0165} & \predscore{0.4768}{0.0072} \\
        $E_{\mathrm{mean}}$ / LR & \ensuremath{0.3496} & \ensuremath{0.7604} & \ensuremath{0.3120} & \ensuremath{0.4740} \\
        $E_{\mathrm{mean}}$ / LGBM & \predscore{0.3112}{0.0263} & \predscore{0.7517}{0.0039} & \predscore{0.3273}{0.0058} & \predscore{0.4634}{0.0079} \\
        $T_{4096}$ / LR & \ensuremath{0.2255} & \ensuremath{0.7049} & \ensuremath{0.3056} & \ensuremath{0.4120} \\
        $T_{4096}$ / LGBM & \predscore{0.2131}{0.0079} & \predscore{0.7066}{0.0028} & \predscore{0.3132}{0.0099} & \predscore{0.4110}{0.0033} \\
        $T_{4096}{+}E_{\mathrm{CLS}}$ / LR & \ensuremath{0.3283} & \ensuremath{0.7598} & \ensuremath{0.3148} & \ensuremath{0.4676} \\
        $T_{4096}{+}E_{\mathrm{CLS}}$ / LGBM & \predscore{0.3416}{0.0108} & \predscore{0.7629}{0.0021} & \predscore{0.3381}{0.0052} & \predscore{0.4809}{0.0032} \\
        $C{+}E_{\mathrm{CLS}}$ / LR & \ensuremath{0.4291} & \ensuremath{0.7805} & \ensuremath{0.3339} & \ensuremath{0.5145} \\
        $C{+}E_{\mathrm{CLS}}$ / LGBM & \predscore{0.4414}{0.0128} & \predscore{0.7878}{0.0015} & \predscore{0.3479}{0.0063} & \predscore{0.5257}{0.0039} \\
        $C{+}E_{\mathrm{CLS}}$ / XGB & \predscore{0.4554}{0.0102} & \predscore{0.7886}{0.0017} & \predscore{0.3485}{0.0069} & \predscore{0.5308}{0.0048} \\
        $C{+}T_{4096}$ / LR & \ensuremath{0.2396} & \ensuremath{0.7191} & \ensuremath{0.3060} & \ensuremath{0.4216} \\
        $C{+}T_{4096}$ / LGBM & \predscore{0.3513}{0.0075} & \predscore{0.7768}{0.0015} & \predscore{0.3317}{0.0102} & \predscore{0.4866}{0.0048} \\
        $C{+}T_{4096}{+}E_{\mathrm{CLS}}$ / LR & \ensuremath{0.3078} & \ensuremath{0.7638} & \ensuremath{0.3165} & \ensuremath{0.4627} \\
        $C{+}T_{4096}{+}E_{\mathrm{CLS}}$ / LGBM & \predscore{0.3667}{0.0139} & \predscore{0.7865}{0.0020} & \predscore{0.3288}{0.0112} & \predscore{0.4940}{0.0042} \\
        \addlinespace[2pt]
        \multicolumn{5}{@{}l}{\textit{Fine-tuned, task-specific (ST)}} \\
        EHR-BERT (ST) & \predscore{0.4564}{0.0064} & \predscore{\mathbf{0.8072}}{0.0006} & \predscore{\underline{0.3504}}{0.0058} & \predscore{\underline{0.5380}}{0.0034} \\
        GLoC-EHR (ST) & \predscore{\mathbf{0.4667}}{0.0123} & \predscore{\underline{0.8038}}{0.0035} & \predscore{\mathbf{0.3778}}{0.0185} & \predscore{\mathbf{0.5494}}{0.0066} \\
        \addlinespace[2pt]
        \multicolumn{5}{@{}l}{\textit{Fine-tuned, multi-task (MT)}} \\
        EHR-BERT (MT) & \predscore{0.4556}{0.0114} & \predscore{0.8009}{0.0027} & \predscore{0.3372}{0.0063} & \predscore{0.5312}{0.0062} \\
        GLoC-EHR (MT) & \predscore{0.4301}{0.0110} & \predscore{0.7952}{0.0057} & \predscore{0.3455}{0.0238} & \predscore{0.5236}{0.0088} \\
        GLoC-EHR (Reasoning) & \predscore{0.3638}{0.0114} & \predscore{0.7498}{0.0011} & \predscore{0.3320}{0.0021} & \predscore{0.4819}{0.0045} \\
        \bottomrule
    \end{tabular*}
    \endgroup
\end{table}

\subsection{Alignment and interface controls}
\label{app:controls}

Table~\ref{tab:controls} reports two single-run controls. \emph{Without
alignment}, the GLoC-EHR (MT) architecture is built on the raw backbone with the
pretrained EHR encoder but randomly initialized resampler, cross-attention,
projection, and selector, and these modules are trained on the three tasks with
the backbone frozen. Selection is soft, because a randomly initialized selector
cannot be trained under hard top-$k$ selection, and the learning rate is chosen
on validation macro AUROC. AUROC is lower than with alignment on all three
tasks (macro 0.862 vs.~0.873), with similar AUPRC (0.535 vs.~0.524). The
\emph{LoRA} control fine-tunes Qwen3-1.7B (rank 128, $\alpha=256$) on serialized
records to answer Yes or No directly, with one learning rate ($10^{-4}$) and one seed. Validation macro
AUROC is computed every 200 steps on a validation subset, training stops after
three evaluations without improvement, and the step-1{,}800 checkpoint is
evaluated. It reaches a macro AUROC of 0.853, below GLoC-EHR (MT) and the
count-based LightGBM (0.864).

\begin{table}[!htbp]
    \centering
    \caption{Single-run controls on the MIMIC-IV test set: AUROC per task and three-task macro AUROC/AUPRC.
    GLoC-EHR (MT) is the mean $\pm$ SD over five seeds from Table~\ref{tab:outcome_prediction}. Without alignment: the same
    architecture trained from randomly initialized interface modules; Qwen3-1.7B + LoRA: the backbone fine-tuned on the
    serialized record. Both are single runs (Appendix~\ref{app:controls}).}
    \label{tab:controls}
    \begingroup
    \footnotesize
    \setlength{\tabcolsep}{3pt}
    \newcommand{\predscore}[2]{\ensuremath{#1{\scriptstyle\,\pm\,#2}}}
    \begin{tabular*}{\linewidth}{@{\extracolsep{\fill}}lccccc@{}}
        \toprule
        & \multicolumn{3}{c}{AUROC $\uparrow$} & \multicolumn{2}{c}{Macro $\uparrow$} \\
        \cmidrule(lr){2-4}\cmidrule(l){5-6}
        Model & Mortality & Long LOS & Readmission & AUROC & AUPRC \\
        \midrule
        GLoC-EHR (MT) & \predscore{0.9545}{0.0013} & \predscore{0.8671}{0.0030} & \predscore{0.7976}{0.0052} & \predscore{0.8731}{0.0030} & \predscore{0.5236}{0.0088} \\
        \quad without alignment & 0.9470 & 0.8578 & 0.7817 & 0.8622 & 0.5354 \\
        Qwen3-1.7B + LoRA & 0.9414 & 0.8411 & 0.7762 & 0.8529 & 0.4990 \\
        \bottomrule
    \end{tabular*}
    \endgroup
\end{table}

\subsection{Reasoning evaluation and answer coverage}
\label{app:reasoning_snapshot}

GLoC-EHR (Reasoning) is the final checkpoint of a single GRPO training run on
the three tasks (step 3{,}216). For each test case we sample five chains
($T{=}0.8$, top-$p$ 0.95, top-$k$ 20, at most 1{,}536 reasoning tokens). The
score of a chain is the Yes/No answer probability conditioned on the prompt and
the generated reasoning prefix. Each chain index is treated as one sampling
run: AUROC and AUPRC are computed separately for each of the five
runs, the macro score averages the three tasks within a run, and we report the
mean and SD over runs, as for training seeds. A chain that does not reach the
answer anchor is excluded from its own run. Because all runs come from one
training run, the SD reflects decoding rather than training variability.
Table~\ref{tab:reasoning_coverage} contrasts anchor coverage with the
zero-shot CoT baselines, and Table~\ref{tab:reasoning_sensitivity} compares
this protocol with scoring chains without an anchor as 0.5 and with averaging
the five chains of each case. These outcome metrics do not establish evidence
support or reasoning faithfulness, which are evaluated separately. Confidence
intervals in the evidence and transfer analyses are two-sided 95\% paired case
bootstrap intervals \citep{efron1993bootstrap}: test cases are resampled with
replacement, and the compared models are scored on the same resamples. Across the
test set, 99.6\% of the 105{,}870 chains reach the answer anchor, and 13 of the
21{,}174 cases have no chain that does. Scoring chains without an anchor as 0.5
changes the macro AUROC by less than 0.001, and averaging the five chains of each
case raises it only slightly, from 0.8656 to 0.8668.

\begin{table}[!htbp]
    \centering
    \caption{Answer-anchor coverage of generated chains (cases reaching a parseable answer; $n$ = 6{,}804 / 6{,}804 / 7{,}566; GLoC-EHR (Reasoning) counts all five chains per case). For the zero-shot baselines, remaining cases are scored by forced closure.}
    \label{tab:reasoning_coverage}
    \small
    \begin{tabular}{lrrrr}
        \toprule
        Model & Mortality & Long LOS & Readmission & Overall (\%) \\
        \midrule
        GLoC-EHR (Reasoning) & 33{,}966 & 33{,}883 & 37{,}623 & 99.62 \\
        Qwen3-1.7B (CoT) & 6{,}799 & 6{,}799 & 7{,}562 & 99.93 \\
        Qwen3-1.7B (CoT, our prompt) & 6{,}380 & 6{,}529 & 7{,}139 & 94.68 \\
        Llama-3.3-70B (CoT) & 6{,}804 & 6{,}804 & 7{,}566 & 100.00 \\
        Llama-3.3-70B (CoT, our prompt) & 6{,}804 & 6{,}804 & 7{,}566 & 100.00 \\
        \bottomrule
    \end{tabular}
\end{table}

\begin{table}[!htbp]
    \centering
    \caption{Scoring sensitivities of GLoC-EHR (Reasoning); unweighted three-task macro. Single-chain rows are means over the five chains; $n$ counts cases.}
    \label{tab:reasoning_sensitivity}
    \small
    \begin{tabular}{lrrr}
        \toprule
        Evaluation variant & $n$ & Macro AUROC & Macro AUPRC \\
        \midrule
        Single chains{,} valid anchors only (reported) & 21{,}094 & 0.8656 & 0.4819 \\
        Single chains{,} no anchor $\mapsto 0.5$ & 21{,}174 & 0.8654 & 0.4814 \\
        Average of five chains{,} valid anchors & 21{,}161 & 0.8668 & 0.4897 \\
        Average of five chains{,} forced closure & 21{,}174 & 0.8665 & 0.4897 \\
        \bottomrule
    \end{tabular}
\end{table}

\paragraph{Evidence of the zero-shot CoT baselines.}
Table~\ref{tab:llm_evidence} scores the evidence bullets of the ``CoT, our
prompt'' baselines with the reward matcher at $\tau_0{=}0.55$, as in
Table~\ref{tab:reasoning_ablation}, over chains that reach the answer anchor.
These models read the record serialized as concept names, with measurements
described in plain language (for example, ``Lactate -- high, above the
reference range''), so a bullet can repeat a line of the prompt: 94.1\% of the
bullets of Llama-3.3-70B and 58.4\% of those of Qwen3-1.7B match a recorded
concept exactly or by containment. GLoC-EHR (Reasoning) receives the record only
through its EHR memories. Qwen3-1.7B, the backbone of GLoC-EHR, reaches a
slightly higher support precision than GLoC-EHR (0.452 vs.~0.424) but cites
fewer distinct supported findings (3.88 vs.~4.16), and Llama-3.3-70B reaches
0.752; both, however, answer near chance (macro AUROC 0.509 and 0.549).
Stricter matching (Figure~\ref{fig:evidence_curves})
is not compared across these systems, because the IDF-weighted overlap also
penalizes wording that the serialization adds around a concept name, such as
``below the reference range'', which GLoC-EHR never sees.

\begin{table}[!htbp]
    \centering
    \caption{Evidence of GLoC-EHR (Reasoning) and the zero-shot ``CoT, our prompt'' baselines on the MIMIC-IV test
    set, scored with the reward matcher ($\tau_0{=}0.55$) over chains that reach the answer anchor. GLoC-EHR
    (Reasoning): mean $\pm$ SD over five sampled chains, as in Table~\ref{tab:reasoning_ablation}. The zero-shot models
    read the serialized record; GLoC-EHR receives it only through its EHR memories. Macro AUROC is taken from
    Table~\ref{tab:outcome_prediction_full}.}
    \label{tab:llm_evidence}
    \begingroup
    \small
    \setlength{\tabcolsep}{3pt}
    \newcommand{\predscore}[2]{\ensuremath{#1{\scriptstyle\,\pm\,#2}}}
    \begin{tabular*}{\linewidth}{@{\extracolsep{\fill}}lccccc@{}}
        \toprule
        & & Support & Unsupported & Distinct & Macro \\
        Model & Bullets & precision $\uparrow$ & rate $\downarrow$ & supported $\uparrow$ & AUROC $\uparrow$ \\
        \midrule
        GLoC-EHR (Reasoning) & 7.83 & \predscore{0.424}{0.001} & \predscore{0.440}{0.001} & \predscore{4.16}{0.01} & \predscore{0.8656}{0.0008} \\
        Qwen3-1.7B (CoT, our prompt) & 6.69 & 0.452 & 0.420 & 3.88 & 0.5087 \\
        Llama-3.3-70B (CoT, our prompt) & 6.85 & 0.752 & 0.060 & 6.33 & 0.5491 \\
        \bottomrule
    \end{tabular*}
    \endgroup
\end{table}

\subsection{Evidence during GRPO training}
\label{app:training_dynamics}

\begin{figure}[t]
    \centering
    \includegraphics[width=\linewidth]{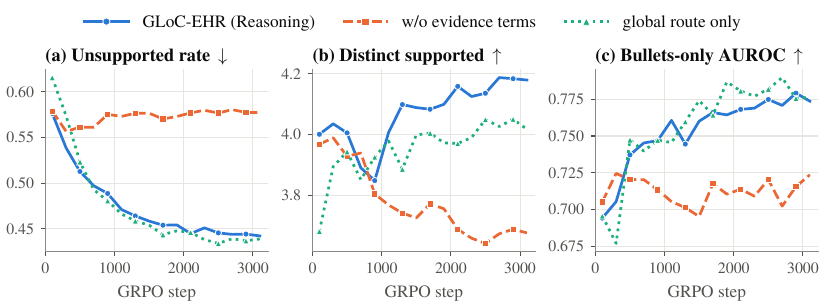}
    \caption{Evidence quality over GRPO training (training rollouts of all ranks, 200-step bins): (a) share of
    evidence bullets that match no event in the input window, (b) distinct supported findings per response, and
    (c) AUROC of the answer predicted from the bullets alone with the EHR memories switched off.}
    \label{fig:training_dynamics}
\end{figure}

Figure~\ref{fig:training_dynamics} tracks the evidence metrics of
Section~\ref{sec:evidence} over GRPO training for GLoC-EHR (Reasoning) and for
the two variants trained without the evidence terms (the evidence reward and the
evidence-sufficiency loss) or without the local route, using the training
rollouts of all four ranks in 200-step bins.
Without the evidence terms, support stays near its SFT level. At the end of
training (steps 3{,}000--3{,}199), 57.7\% of its bullets are unsupported
vs.~44.2\% for GLoC-EHR, it cites fewer distinct supported findings (3.68
vs.~4.18), and its bullets predict the outcome less well (bullets-only AUROC
0.724 vs.~0.774), at similar balanced accuracy (0.734 vs.~0.740). The
global-route-only variant, trained from its own SFT checkpoint, ends with a
similar unsupported rate (0.439 vs.~0.442) and bullets-only AUROC (0.776
vs.~0.774) but fewer distinct supported findings (4.02 vs.~4.18). Prompts are
sampled with replacement.

\subsection{Zero-shot transfer to unseen questions}
\label{app:transfer}

We ask GLoC-EHR (Reasoning) two laboratory questions that it never saw in
training: whether the next serum sodium will be below 135\,mmol/L and whether
the next platelet count will be below 150\,K/$\mu$L, given the admission's
trajectory up to that measurement. Only the question in the prompt changes. The
cases come from MIMIC-IV admissions of held-out patients (none in the training
split; 1{,}871 and 1{,}894 patients). In each admission, the first measurement
below the threshold is taken as a positive and one randomly chosen measurement
at or above it as a negative; negatives are matched to positives on sequence
length, and 750 positives and 1{,}500 negatives are drawn per question.
Table~\ref{tab:transfer} compares one sampled chain of GLoC-EHR (Reasoning)
(chains without an answer, 2.2\%, are excluded) with zero-shot Qwen3-1.7B
reading the serialized record, prompted as in Table~\ref{tab:outcome_prediction}.
Neither model is trained on these questions. GLoC-EHR (Reasoning) is higher by
0.100 [0.068, 0.131] and 0.104 [0.071, 0.137] AUROC (paired case bootstrap).

\begin{table}[t]
    \centering
    \caption{Zero-shot answers to two laboratory questions absent from training (held-out MIMIC-IV admissions;
    750 positives and 1{,}500 length-matched negatives per question, so AUPRC is 0.33 by chance). GLoC-EHR
    (Reasoning) scores one sampled chain per case.}
    \label{tab:transfer}
    \begingroup
    \footnotesize
    \begin{tabular}{lcccc}
        \toprule
        & \multicolumn{2}{c}{Next sodium $<$ 135 mmol/L} & \multicolumn{2}{c}{Next platelet count $<$ 150 K/$\mu$L} \\
        \cmidrule(lr){2-3}\cmidrule(l){4-5}
        Method & AUROC & AUPRC & AUROC & AUPRC \\
        \midrule
        GLoC-EHR (Reasoning) & 0.657 & 0.500 & 0.629 & 0.499 \\
        Qwen3-1.7B, zero-shot & 0.556 & 0.410 & 0.525 & 0.344 \\
        \bottomrule
    \end{tabular}
    \endgroup
\end{table}

\end{document}